\documentclass[sn-mathphys,iicol]{sn-jnl}
\usepackage{makecell}
\usepackage{tablefootnote}
\usepackage{float}
\usepackage{relsize}
\usepackage{array, threeparttable, booktabs,  caption}
\usepackage{dblfloatfix}

\jyear{2023}%

\theoremstyle{thmstyleone}%
\theoremstyle{thmstyletwo}%

\theoremstyle{thmstylethree}%

\begin{document}

\title[Face2PPG]{\vspace{-2cm}Face2PPG: An unsupervised pipeline for blood volume pulse extraction from faces}


\author*[1]{\fnm{Constantino} \sur{Álvarez Casado}}\email{constantino.alvarezcasado@oulu.fi}

\author[1,2]{\fnm{Miguel} \sur{Bordallo López}}\email{miguel.bordallo@oulu.fi}

\affil*[1]{\orgdiv{Center for Machine Vision and Signal Analysis}, \orgname{University of Oulu}, \orgaddress{ \country{Finland}}}

\affil[2]{\orgdiv{Cognitive technologies for intelligence}, \orgname{VTT Technical Research Centre of Finland Ltd}, \orgaddress{  \city{Oulu}, \country{Finland}}}


\abstract{Photoplethysmography (PPG) signals have become a key technology in many fields, such as medicine, well-being, or sports. Our work proposes a set of pipelines to extract remote PPG signals (rPPG) from the face robustly, reliably, and configurable. We identify and evaluate the possible choices in the critical steps of unsupervised rPPG methodologies. We assess a state-of-the-art processing pipeline in six different datasets, incorporating important corrections in the methodology that ensure reproducible and fair comparisons. In addition, we extend the pipeline by proposing three novel ideas; 1) a new method to stabilize the detected face based on a rigid mesh normalization; 2) a new method to dynamically select the different regions in the face that provide the best raw signals, and 3) a new RGB to rPPG transformation method, called Orthogonal Matrix Image Transformation (OMIT) based on QR decomposition, that increases robustness against compression artifacts. We show that all three changes introduce noticeable improvements in retrieving rPPG signals from faces, obtaining state-of-the-art results compared with unsupervised, non-learning-based methodologies and, in some databases, very close to supervised, learning-based methods. We perform a comparative study to quantify the contribution of each proposed idea. In addition, we depict a series of observations that could help in future implementations.}




\maketitle

%
%

\section{Introduction}
\label{sec:introduction}

Photoplethysmography (PPG) signals have become a key technology in many fields, such as medicine, well-being, or sports. The technology utilizes a light source and a photodetector to measure the blood volume pulse (BVP) as light variations in skin tissues \cite{PPG2007Tech}. In medicine, PPG analysis is a basic and common tool in healthcare services to monitor vital signs such as heart rate (HR) or oxygen saturation (SpO\textsubscript{2}) \cite{CurrentStatePPGs4Health}. In well-being, it became increasingly important thanks to the success of wearable devices that analyze sleep disorders \cite{PPG4SleepControl2017}, cardiovascular diseases \cite{OuraCVDs2020Kinnunen}, or detection of stress and meditation \cite{MeditationDetection2021Alvarez}. In sports, PPG analysis became an important tool to improve the intrinsic and extrinsic athletes' performance \cite{PPGMonitoring2019SportPerformance}. 

Remote PPG (rPPG) imaging is a contactless version of this technology that uses video cameras, usually consumer-grade RGB or near-infrared cameras, and ambient light sources. It works by recording a subject's face or body parts with visible skin areas and analyzing the subtle color variations or motion changes in skin regions \cite{rPPGFundaments2015}\cite{RecentReviewRPPGMethods2021}. The remote PPG technique allows for non-invasive evaluation and monitoring of users in services, such as healthcare. Hence, the technology could offer significant advantages compared to contact-based devices if it becomes reliable \cite{RemoteMonitoringHealth2019Review}.

\begin{figure*}[htb!]
  \begin{center}
    \includegraphics*[width=0.99\textwidth]{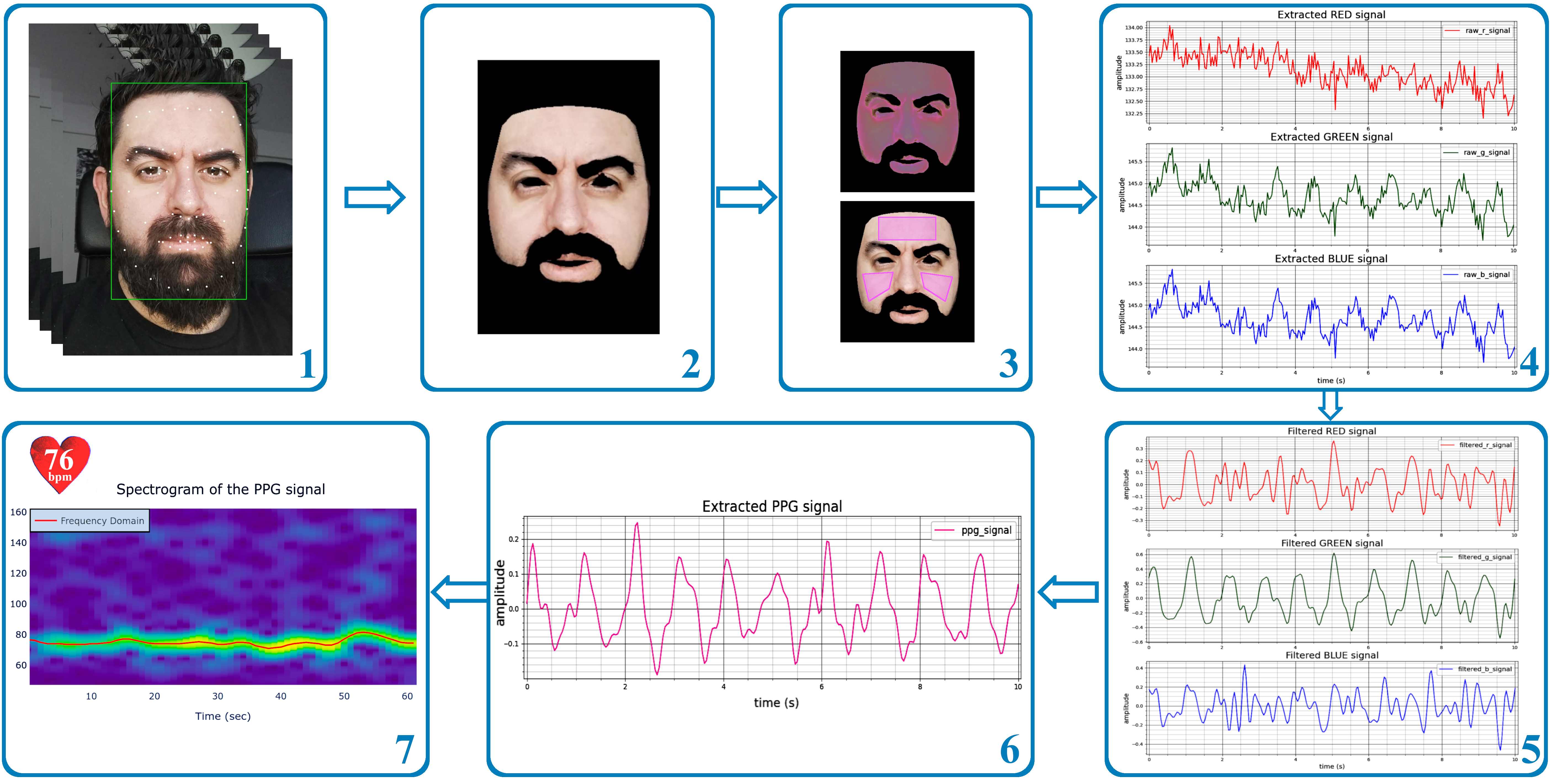}
  \end{center}

    \caption{Typical unsupervised based methodology for remote photoplethysmographic (PPG) imaging using a RGB camera. It comprises several steps: 1) Face detection and alignment 2) Skin segmentation 3) ROI selection 4) Extraction of the raw signals from ROIs 5) Filtered signals 6) RGB to PPG transformation 7) Spectral analysis and post-processing.}
  \label{fig:PPGProcess}
\end{figure*}

Current approaches for recovering physiological signals from videos are mainly unsupervised non-learning-based methods and supervised deep learning approaches \cite{RecentReviewRPPGMethods2021}. Deep learning-based methods propose end-to-end solutions utilizing training datasets. In contrast, unsupervised non-learning-based (sometimes named "Traditional") methods employ computer vision and signal processing in a structured pipeline, as illustrated in Figure \ref{fig:PPGProcess}.

Most of the unsupervised rPPG methods proposed in the literature focus on recovering PPG signals mostly from static faces, disregarding challenges under real scenarios in real-world applications such as fast face and head movements, extreme light conditions, facial expressions, illumination changes, occlusion, or distance from the camera to the subject.

This article focuses on improving the performance of state-of-the-art rPPG unsupervised, non-learning-based methods, emphasizing all system components. We tackle the improvement of the process by proposing a set of changes across the whole pipeline that result in a noticeable global improvement, performing an extensive evaluation and a framework to recover physiological signals from faces.

%
%

\subsection{Contributions}\label{sec:contributions}

This article depicts different performance problems and challenges to recovering PPG signals from faces reliably. We improve several components of the rPPG pipeline with novel ideas. The main contributions can be summarized as follows:

\begin{itemize}
\setlength\itemsep{0pt}
\setlength\parskip{0pt}
\item We provide a new method to stabilize the movement and facial expression based on a rigid mesh normalization, ensuring that the raw RGB signals are measured from the same facial location regardless of the pose and movement.

\item We provide a new method based on statistical and fractal analysis to dynamically select only the facial regions that supply the best raw signals, discarding those with higher noise or prone to artifacts.

\item We propose a novel rPPG method to transform the RGB signal into a PPG signal based on QR decomposition, named Orthogonal Matrix Image Transformation (OMIT), which proves to be robust to video compression artifacts.

\end{itemize}

To prove the usefulness of our approach, we extensively evaluate a set of rPPG methods with four different pipelines across several datasets. Our experiments include modifications to the original evaluation pipeline to increase the fairness and reproducibility of the comparative results.

%
%

\section{Related work}\label{sec:relatedwork}
In the last few years, rPPG research has progressed from the filtering and simple processing of the variation of the facial skin color to sophisticated multi-step processing pipelines and end-to-end supervised learning methods with dedicated architectures.

\subsection{Unsupervised methods}\label{sec:PPGClassicalMethods}
Unsupervised non-learning-based methods focus on recovering physiological signals by applying computer vision and signal processing techniques as a system with several steps. These methods focus on obtaining the BVP signal by finding skin areas suitable to extract the raw RGB signals, using face detection, tracking, and segmentation techniques. After that, these methods carefully process these raw RGB signals to separate the physiological signals contained in the subtle variations of the skin color from the rest of the information (motion, illumination changes, or facial expressions, among others) by applying filtering and different ways of combining the RGB signals into an rPPG signal. Most of the studies focus mainly on this transformation component using similar approaches and components for the rest of the process \cite{RecentReviewRPPGMethods2021}. 

\cite{GreenMethod2008} proposed the first study on extracting remote PPG signals using an inexpensive consumer-grade RGB camera. The study showed how the green channel of the camera contains rich enough, significant information to recover signals such as the heart pulse. \cite{PohMethod2011} proposed the recovery of physiological signals by applying the blind source separation (BSS) technique to remove the noise. Concretely, they used Independent Component Analysis (ICA) to uncover the independent source signals. Similar to this work, \cite{PCAMethod2011} proposed Principal Component Analysis (PCA) to reduce the computational complexity in comparison to Independent Component Analysis with similar accuracy performance. \cite{CHROMMethod2013} proposed a chrominance-based method (CHROM) to separate the specular reflection component from the diffuse reflection component, which contains pulsatile physiological signals, both reflected from the skin and based on the dichromatic reflection model. \cite{PVBMethod2014} define a Blood-Volume Pulse (PBV) vector that contains the signature of the specific blood volume variations in the skin, removing noise and motion artifacts. In the same year, \cite{LiMethod2014} focused on removing the human motions and artifacts from the RGB signals by applying Normalized Least Mean Square (NLMS) adaptive filter. They perform this rectification step by assuming both the face ROI and the background as Lambertian models that share the same light source. \cite{2SRMethod2016} proposed a data-driven algorithm by creating a subspace of skin pixels and the computation of the temporal rotation angle of the computed subspace between subsequent frames to extract the heart rate pulse. \cite{Lab2016SNR} proposed the CIELab color space (LAB) transformation as a more robust color space to extract pulse rate signals due to the high separation between the intensity and chromaticity components, less sensitive to human body movements. The study also demonstrates that the \textit{a} channel has a better signal-to-noise ratio (SNR) than the green channel in RGB color space. \cite{POSMethod2017} proposed a new plane-orthogonal-to-skin (POS) algorithm that finds pulsatile signals in an RGB normalized space orthogonal to the skin tone. \cite{LGIMethod2018} proposed the Local Group Invariance (LGI) method, a stochastic representation of the pulse signal based on a model that leverages the local invariance of the heart rate as a quasi-periodical process dynamics and obtained by recursive inference to remove extrinsic factors such as head motion and lightness variations. \cite{FaceRPPG2020Gudi} proposed an unsupervised method with an emphasis on motion suppression and novel filtering based on the head orientations (FaceRPPG). Recently, deep learning methods have been developed that do not rely on reference signals, which could be considered unsupervised \cite{gideon2021way}\cite{sun2022contrast}. However, they remain specifically tailored to each dataset and its unique characteristics.

Unsupervised methods offer a significant advantage in that they do not necessitate specific training data, allowing for better generalization across different datasets and measurement setups. These methods focus on measuring the BVP signal as it manifests in various facial regions, ensuring adaptability to diverse scenarios. However, given the absence of a learning component, the performance of unsupervised methods can be influenced by several factors: sensitivity to noise and artifacts, dependency on user-defined parameters (many unsupervised methods rely on user-defined parameters for their performance, making them less adaptable to diverse scenarios and subjects), and limited adaptability to varying skin types and lighting conditions (previous unsupervised methods may not be well-suited for handling diverse skin types and lighting conditions, which can result in suboptimal rPPG signal extraction).

Unsupervised non-learning-based method's efforts have focused on finding suitable ways of transforming noisy RGB signals into reliable PPGs. In contrast, the impact of other system components, such as face detection and tracking, has been mainly disregarded. Our contribution addresses the unsupervised rPPG process as a system with multiple components that can be improved separately.

\subsection{Supervised methods}\label{sec:DeepLearningMethods}

\textit{Deep Neural Networks} (DL) and especially \textit{Convolutional Neural Networks} (CNN) approaches have gained attention and become popular tools in computer vision and signal processing tasks, including healthcare-related tasks. Before the advent of deep-learning based methods, there were two primary approaches to estimate heart rate using machine learning methods, including support-vector regression \cite{Hsu2014SVMHR} and adaptive hidden Markov models \cite{Fan2015BayesHR}. Since 2018, supervised deep learning-based methods to compute HR or other vital signs started to arise increasingly in the literature. Among the most relevant learning-based remote PPG methods, \cite{HRCNNMethod2018} proposed a two-step convolutional neural network to estimate a heart rate value from a sequence of facial images. HR-CNN is a trained end-to-end network composed of two components, an Extractor and an HR Estimator. The same year, \cite{DeepPhysMethod2018} proposed DeepPhys, another end-to-end solution based on a deep convolutional network that estimates HR and breathing rate (BR). The approach performs a motion analysis based on attention mechanisms and a skin reflection model using appearance information to extract the physiological signals. \cite{RhythmNet2019} proposed RhythmNet, an end-to-end solution based on spatial-temporal mapping to represent the HR signals in videos. The approach also exploits temporal relationships of adjacent HR estimations to perform continuous heart rate measurements. The same year, \cite{ZitongMethod2019} proposed a two-stage end-to-end solution. The first part of the network, named STVEN, is a Spatio-Temporal Video Enhancement Network to improve the quality of highly compressed videos. The second part of the approach, called rPPGNet, is the 3D-CNN network that recovers the rPPG signals from the enhanced videos. The authors claim that the proposed rPPGNet produces rich rPPG signals with curve shapes and peak locations. \cite{AutoHR2020Zitong} further proposed another end-to-end approach for remote HR measurement based on Neural Architecture Search (NAS). AutoHR is comprised of three ideas: a first stage that discovers the best topology network to extract the physiological signals based on Temporal Difference Convolution (TDC); a hybrid loss function based on temporal and frequency constraints; and Spatio-temporal data augmentation strategies to improve the learning stage. The same year, \cite{MetaRPPG2020} proposed a transductive meta-learner based on an LSTM estimator and Synthetic Gradient Generator that adjusts network weights in a self-supervised manner. Recently, \cite{PulseGAN2021} proposed a two-stage hybrid method called PulseGAN. It starts with the unsupervised extraction of noisy PPG signals using CHROM as a first stage, followed by a generative adversarial network (GAN) that generates realistic rPPG pulse signals from the signals recovered in the first stage. In 2021, \cite{ANDrPPG2022} proposed a novel denoising-rPPG method called AND-rPPG, based on the utilization of Action Units (AUs) and Temporal Convolutional Networks (TCNs) for denoising temporal signals to mitigate facial expression noises effectively.

Supervised rPPG methods, primarily leveraging deep learning, exhibit exceptional accuracy by employing end-to-end solutions that aim to extract physiological signals or heart rate values directly from video data. Apart from essential preprocessing, these methods necessitate minimal intermediate steps \cite{Survey_DL_RPPG2021_Cheng}. Through observing facial noise patterns, supervised methods attempt to learn reference contact-based PPG ground-truth signals from the finger, culminating in a black box model that recovers physiological signals from video frames without a clear understanding of the underlying mechanisms. This lack of transparency presents significant challenges to the practical application of these signals in critical areas such as healthcare \cite{CNNrPPG_Limitations2020}, as well as in scenarios where the available data for training is limited and prone to anomalies (e.g., for cardiovascular conditions). Furthermore, the dependency of supervised methods on labeled training data raises issues of data acquisition and cost, especially in medical contexts where privacy and ethical concerns are of utmost importance. Supervised methods may also face difficulties in generalizing to new scenarios, subjects, or recording conditions if the training data does not encompass a wide array of variations. Moreover, supervised methods may need substantial computational resources.

%
%

\section{Unsupervised blood volume pulse extraction methodology}\label{sec:BVPMethodology}

Our work uses a standard methodology in unsupervised rPPG approaches to extract the blood volume changes from facial videos and derive essential parameters. We follow a modular pipeline with several components, as depicted in Figure \ref{fig:PPGProcess}. The pipeline is roughly divided into three big blocks: the selection of measuring regions of the face, the extraction of rPPG biosignals from natural variations in color or texture, and the computation of the heart rate or other parameters using the extracted signals. 

The pipeline comprises of 8 main modules sequentially connected:
\begin{enumerate}
\setlength\itemsep{0pt}
\setlength\parskip{0pt}
\item \textbf{Database interface}: Provides interfaces to read data from various public databases, including videos, images, and reference signals.
\item \textbf{Face detection and alignment}: An initial step for detecting and aligning faces in each frame. This process obtains face coordinates and facial points, which are crucial for selecting regions of interest, patches, or segmentation coordinates.
\item \textbf{ROI selection}: step to select the regions of interest of the face based on a color skin segmentation or selection of patches based either on face location or the coordinates of the landmarks.
\item \textbf{RGB extraction}: Retrieves raw signal from a window of multiple RGB frames, computing the mean value of pixels within the mask or patches selected earlier.
\item \textbf{Pre-processing}: Block mainly to filters raw RGB signal to focus on the band of interest. Filtering is essential for accurate BVP signal recovery, as remote PPG signals often exhibit trends and noise. The heart-related signal lies between 0.75 and 4.0 Hz.
\item \textbf{RGB to PPG transformation (rPPG)}: crucial and core step in remote PPG methods as it converts skin color variations into physiological signals. This step involves transforming the RGB signal into a PPG signal using a transformation method that combines the RGB channels to generate a pulse signal.
\item \textbf{Frequency analysis}: spectrum analysis of the rPPG and reference ground-truth (BVP or ECG) signals to estimate the heart rate. 
\item \textbf{Evaluation}: it includes an error and statistical analysis to compare the estimated heart rate from the rPPG and the estimated heart rate from the ground-truth signals.
\end{enumerate}

\vspace{-3mm}
\subsection{Baseline pipeline}
In our work, we start from an open-source framework for the evaluation of remote PPG methods, implemented in Python and called \textit{pyVHR} (short for Python tool for Virtual Heart Rate)\cite{pyVHR2020}. This framework includes an extensible interface to integrate several datasets, multiple methods, choices for each processing step, and extensive assessment and visualization tools. We use version 0.0.4 of the PyVHR framework, which we will refer to as \textit{Baseline} pipeline onward.

\section{Face2PPG pipelines}
\label{sec:Face2PPGPipelines}

 The \textit{Baseline} pipeline presents a few shortcomings that might result in inaccurate or unfair assessments. We incrementally improve it by introducing several changes in multiple steps, and propose three new versions that we name \textit{Improved}, \textit{Normalized} and \textit{Multi-region} pipelines.

These pipelines focus on handling the face in unconstrained conditions since it is one of the critical parts of extracting remote photoplethysmograms. It has to be noted that most of the unsupervised approaches have focused mainly on developing RGB to PPG conversion methods (sometimes called just rPPG methods), but they have not paid much attention to other steps of the pipeline. This article emphasizes the importance of every step when extracting and evaluating remote PPG signals from faces. Moreover, we have devised a novel method, Orthogonal Matrix Image Transformation (OMIT), which employs QR decomposition to convert raw RGB signals into BVP signals.

\subsection{Improved pipeline}
To mitigate its shortcomings, we modified the \textit{Baseline} pipeline to incorporate a few minor changes that increase reproducibility and enable a fairer comparison of different methods. We name this modified version as \textit{Improved} pipeline. We enumerate and describe these changes as follows:

\textbf{Face detection}:
The \textit{Baseline} pipeline includes two well-know face detectors: one based on convolutional neural networks known as MTCNN \cite{MTCNN2016} and a \textit{Dlib} implementation based on Histogram of Oriented Gradients (HOG) features \cite{dlib09}. We use instead a new deep learning-based face detection method based on a Single Shot Multibox Detection network (SSD) \cite{SSDFaceDetector2015}, implemented in \textit{OpenCV} library. This face detector outperforms the \textit{Baseline} pipeline detectors in terms of accuracy, size of the models, and computational speed \cite{SSDvsMTCNN2021}.

\begin{figure}[ht!]
  \begin{center}
    \includegraphics*[width=0.48\textwidth]{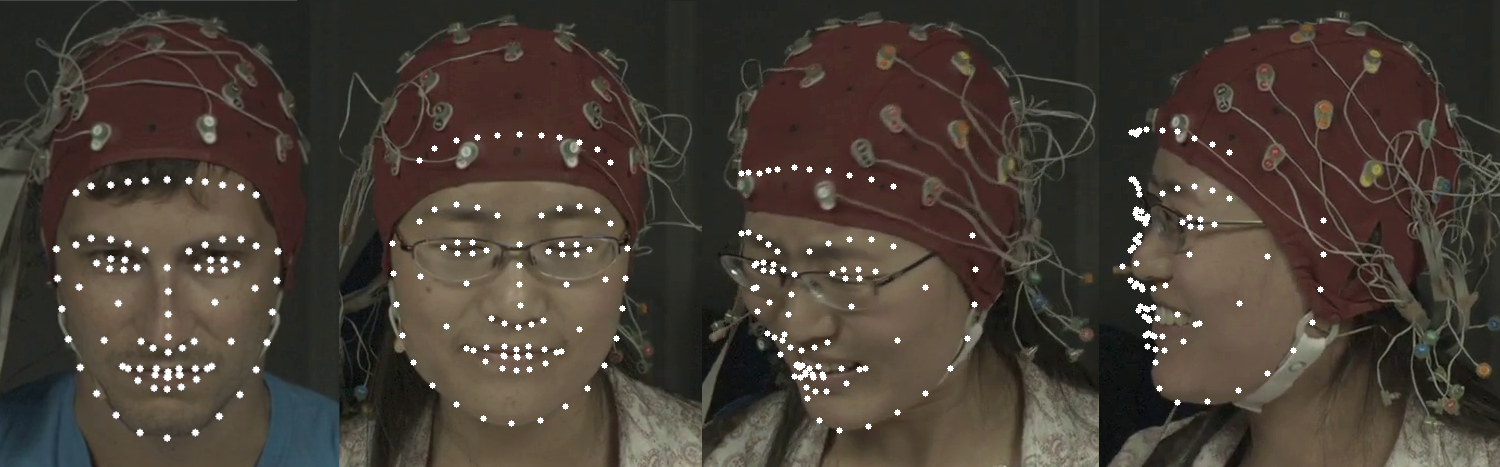}
  \end{center}

  \caption{Facial landmark detection using DAN model under extreme head poses and frontal faces.}
  \label{fig:DAN_Face_Alignment}

\end{figure}

\textbf{Face alignment}:
The \textit{Baseline} pipeline includes two well-know face landmark detectors, the MTCNN detector \cite{MTCNN2016} that computes 5 landmark points in the face (eyes, nose and mouth corners) and the \textit{Dlib} implementation of the ERT  method \cite{ERTFace2014}\cite{dlib09}. We use instead a deep learning approach named DAN (Deep Alignment Network) \cite{DANKowalskiNT17} which gives exceptional performance in terms of accuracy, even in challenging conditions \cite{AlvarezBordallo2021FaceAlignment} as shown in Figure \ref{fig:DAN_Face_Alignment}. For a faster, real-time face alignment, we have added a more accurate, faster, and smoother model for the ERT \textit{Dlib} face landmarks detector \cite{AlvarezBordallo2021FaceAlignment}. These models both infer 68 landmark points defined by the Multi-PIE landmark scheme.

\textbf{Filtering:}
The \textit{Baseline} pipeline only considers a pre-filtering scheme before the RGB to PPG transformation. The pipeline offers three types of filters: detrending (Scipy or Tarvainen methods), bandpass filtering (FIR filter with Hamming window and Butterworth IIR filter), and a Moving average filter (MA) that removes various base noises and motion artifacts of the signals. We have added the possibility of using Kaiser windows when applying FIR filtering. A Kaiser-Besel window maximizes the energy concentration in the main lobe, and it is highly recommended to filter biosignals \cite{KaiserWindow2019}. In addition, we have introduced the possibility of applying also post-filtering, performed after the RGB to PPG conversion, since the literature suggests that some conversion methods perform better this way \cite{Unakafov_2018}.

\textbf{RGB to PPG transformation (rPPG)}:
The \textit{Baseline} pipeline includes several reference methods such as POS \cite{POSMethod2017}, CHROM \cite{CHROMMethod2013}, GREEN\cite{GreenMethod2008}, PCA \cite{PCAMethod2011}, ICA \cite{PohMethod2011}, SSR \cite{2SRMethod2016}, LGI \cite{LGIMethod2018} and PVB \cite{PVBMethod2014}. We have added one method based on selecting the chroma channel \textit{a} after applying a CIE Lab color space transformation. CIE Lab separates the lightness information (channel L) from the chroma information (channels a and b). The chrominance components have a more significant dynamic range than the red, green, and blue channels in RGB color space \cite{Lab2016SNR}. In addition, it correlates with skin color and related parameters and describes better the subtle changes occurring in them \cite{LABSkinColor2020}.

\textbf{Spectral analysis}:
In the original framework, the ground-truth pulse rate is estimated using Short Time Fourier Transform (STFT) when the ground-truth signal is a BVP signal, and R-Peak detection and RR interval analysis when the ground-truth signal is an ECG signal. In the Baseline pipeline, the recovered PPG signals are processed instead of using Welch's spectral density estimation. This mismatch introduces the possibility of unfair evaluation. We modified the pipeline, so the ground-truth BVP signal and the rPPG signal are processed using the same spectral analysis algorithm and similar parameters such as the overlap or FFT length.

\begin{figure}[ht!]
  \vspace{-2mm}
  \begin{center}
    \includegraphics*[width=0.49\textwidth]{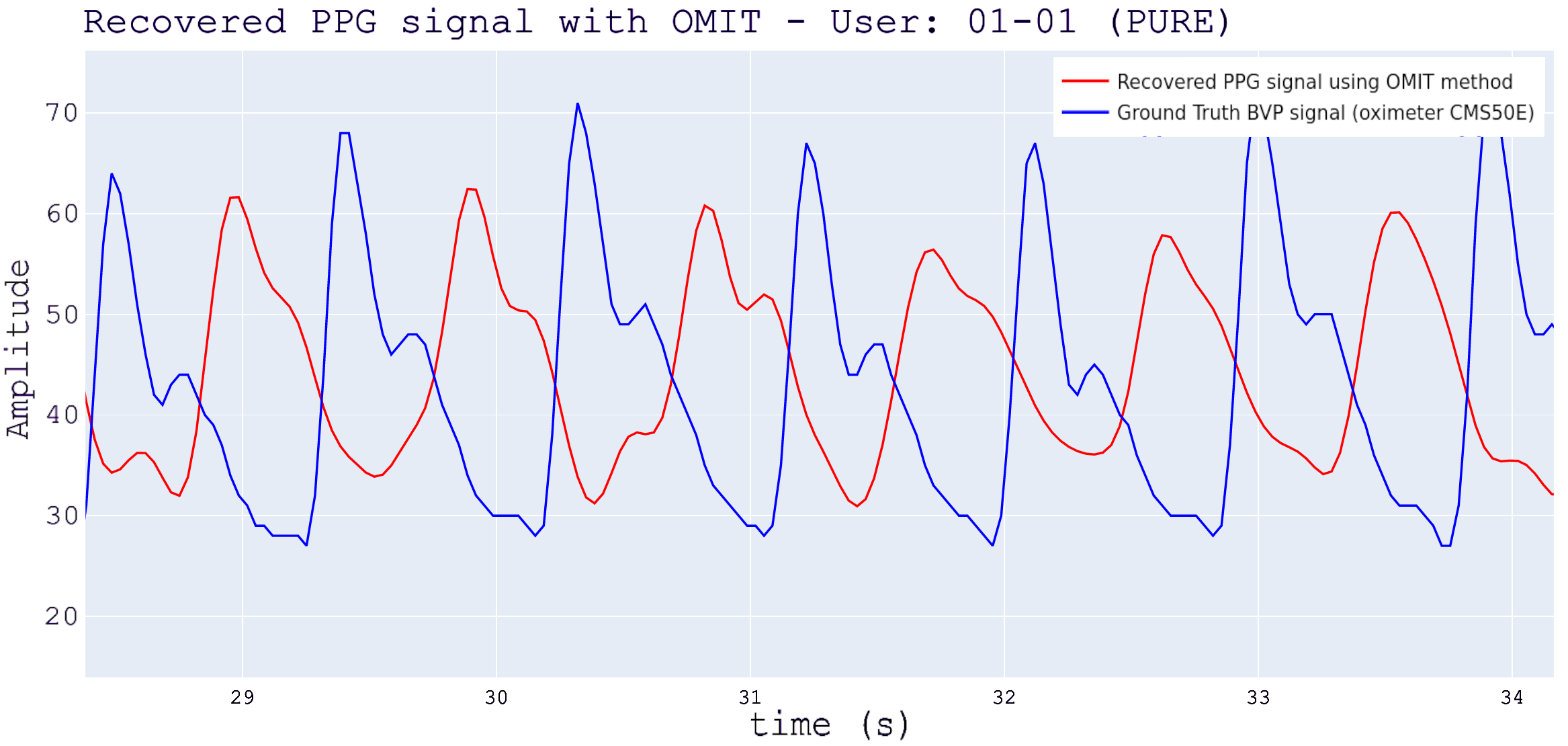}
  \end{center}
  \caption{Delay between reference BVP and remote PPG signals from proposed pipelines, induced by factors like filtering, blood perfusion differences, or camera distance.}
  \label{fig:PURE_Async_BVP}
\end{figure}

\begin{figure}[ht!]
  \vspace{-2mm}
  \begin{center}
    \includegraphics*[width=0.49\textwidth]{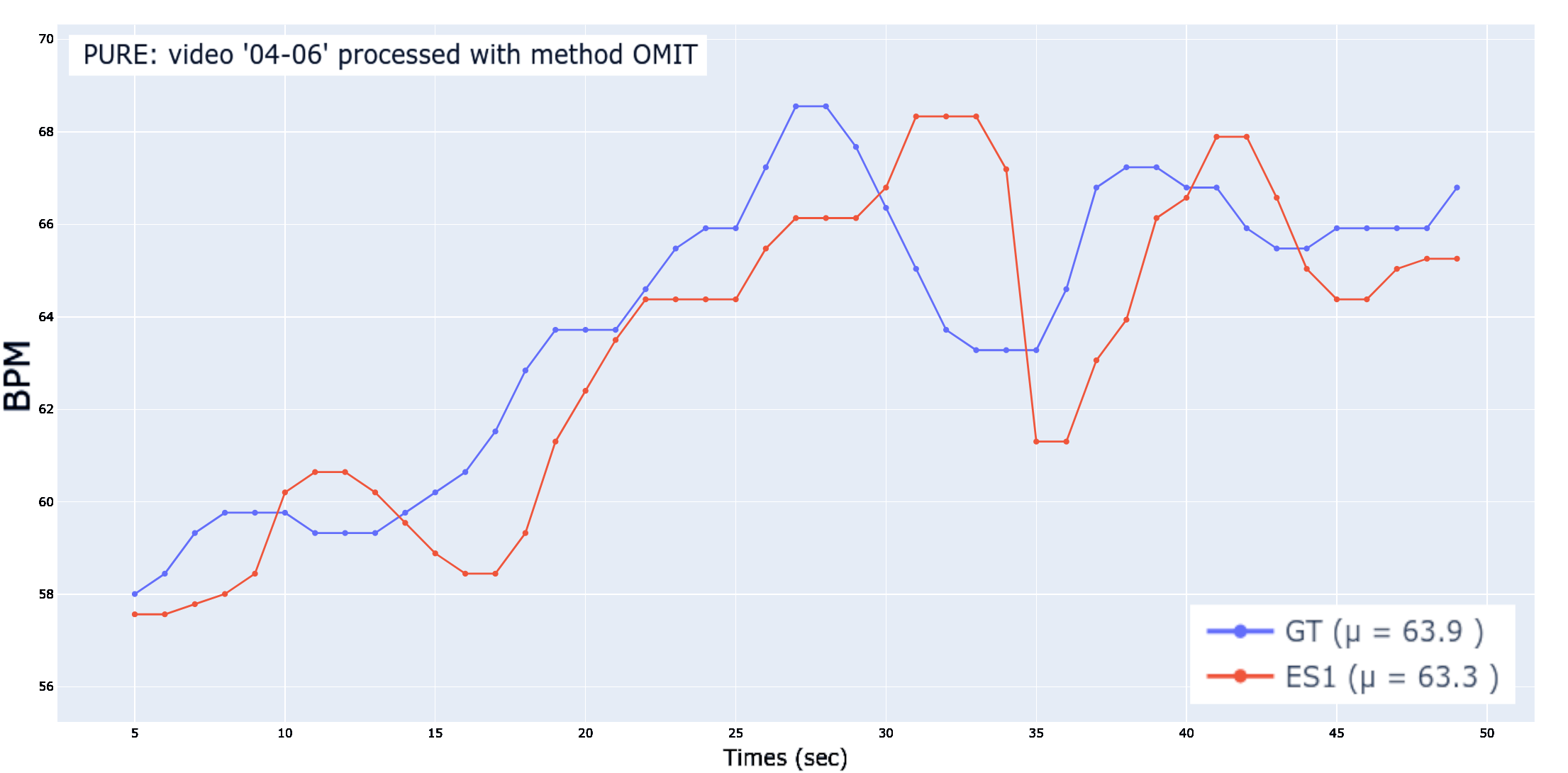}
  \end{center}
  \caption{Estimated HR from the reference BVP signal and the extracted facial rPPG signal using the \textit{Baseline} pipeline. We can observe the asynchrony due to the different signal sources.}
  \label{fig:PURE_Async}
\end{figure}

\textbf{Evaluation}:
It can be expected that the reference BVP (PPG) signals taken in the finger and the rPPG signals extracted from the face are not perfectly synchronized or show the same dynamic range. We show an example in the Figure \ref{fig:PURE_Async_BVP}. We can also observe asynchrony in the heart rate estimation, as shown in Figure \ref{fig:PURE_Async}.

Technical and physiological factors can produce time shifts and morphological differences in the signals. These effects can be attributed to the distance between the optical sensors, the contact force effect of the finger PPG oximeter \cite{PPGFingerPressureChanges}, different filtering parameters \cite{FilteringEffectTime2021}, individual variations among subjects \cite{InaccuracySources2021PPG}, variability in the measurement site \cite{respiratoryPPG2019EffectsSite}, and even blood perfusion differences in different body regions \cite{BloodPerfusionDifferences2018}, among others. We mitigate some of the effects caused by comparing fundamentally different signals by adding a new parameter in the dataset interface that aligns both signals in terms of time and dynamic range, resulting in a fairer estimation of the error between ground-truth HR estimation and rPPG HR estimation. The parameter has been adjusted for each dataset globally using a mostly empirical approach due to the lack of detailed information on the measurement setups. This approach ensures the consistency of signal alignment while maintaining the overall integrity of the data.

\subsection{Normalized pipeline}

Our \textit{Normalized} pipeline introduces two significant changes compared to the previous ones, a segmentation approach based on geometric normalization, and a novel RGB to PPG transformation method, robust to compression artifacts.

\subsubsection{Geometric segmentation and normalization}
One of the critical steps in the non-contact PPG extraction is the process related to skin segmentation, since it is the source to recover the desired physiological signals. Most of the unsupervised methods and pipelines rely on simple thresholding of different color spaces for skin color segmentation \cite{pyVHR2020}, from inefficient fixed RGB segmentation to adaptive HSV segmentation. These pixel-level techniques suffer from generalization due to the variability of skin tones, skin paint, illumination changes, and complex backgrounds. It is not easy to define clear boundaries between the skin and non-skin pixels, mainly due to the variability of the facial regions measured across frames of a single video. Framewise skin segmentation based on neural networks (e.g., uNET) suffers similar problems, sometimes caused by the small number of annotated facial skin masks, resulting in underfitted models \cite{Nanni2023SkinDetection}.  

\begin{figure}[h!]
  \begin{center}
    \includegraphics*[width=0.48\textwidth]{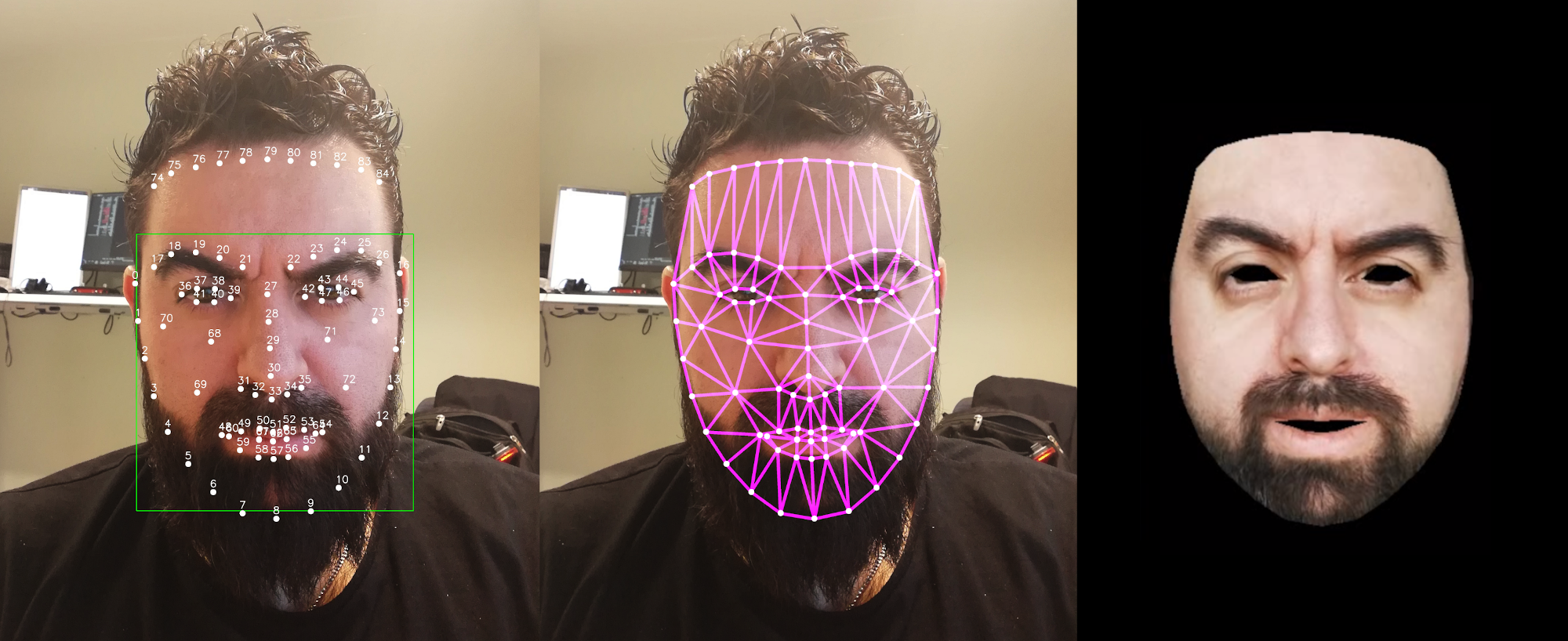}
  \end{center}
  \caption{Face normalization process. Left to right: detected landmark points, fixed triangle face mesh, and normalized face after mapping triangles to fixed coordinates.}
  \label{fig:TriangleMeshTino}
\end{figure}

We propose using a geometrical segmentation scheme that uses fiducial landmark points detected in the face. Although some interframe jittering due to the landmark variability remains and produces changes in the skin mask across the video frames, this is noticeably lower than using skin color segmentation. To perform this segmentation, we have extended the set of landmark points from 68 to 85 landmark points by interpolation and created a fixed facial mesh composed of 131 triangles and fix their coordinates as a typical frontal face, as shown in Figure \ref{fig:TriangleMeshTino}.

Our segmentation approach normalizes the face in each frame by mapping every triangle in the current detected face to the triangles in the normalized shape as shown in Figure \ref{fig:TriangleMeshTino}. This approach generates a spatio-temporal matrix of normalized faces that ensures that we measure the signals in the same facial regions consistently across frames, regardless of the pose and movement.

\vspace{-3mm}
\subsection{Multi-region pipeline}

Extracting only one signal from the whole face or skin mask can result in a very noisy signal. However, the previously described skin segmentation produces a set of facial regions that can be analyzed separately. Due to partial occlusions, extreme head poses, illumination variations, or shades, among others, some of these regions might present very noisy signals with low dynamic ranges. Previous methods have proposed to select fixed patches in the face, where a priori, the blood perfusion should be more observable (e.g., forehead and cheeks). This approach works relatively well when the videos show quasi-static individuals in fixed environments but fails when presented with fast movements or strong face rotations. To mitigate the impact of these challenges, we propose to modify the pipeline by introducing a dedicated block that automatically and dynamically selects those regions that contain the raw signals with higher quality. We name the resulting framework as \textit{Multi-Region Pipeline}.

\subsubsection{Dynamic multi-region selection}

We propose a novel Dynamic Multi-Region Selection (DMRS) method to select the best facial regions dynamically. This approach extracts signals in a fixed set of facial regions and statistically analyzes their quality to choose whether each one of them should contribute to the final rPPG signal or if it should be discarded.

\begin{figure}[ht!]
  \begin{center}
    \includegraphics*[width=0.48\textwidth]{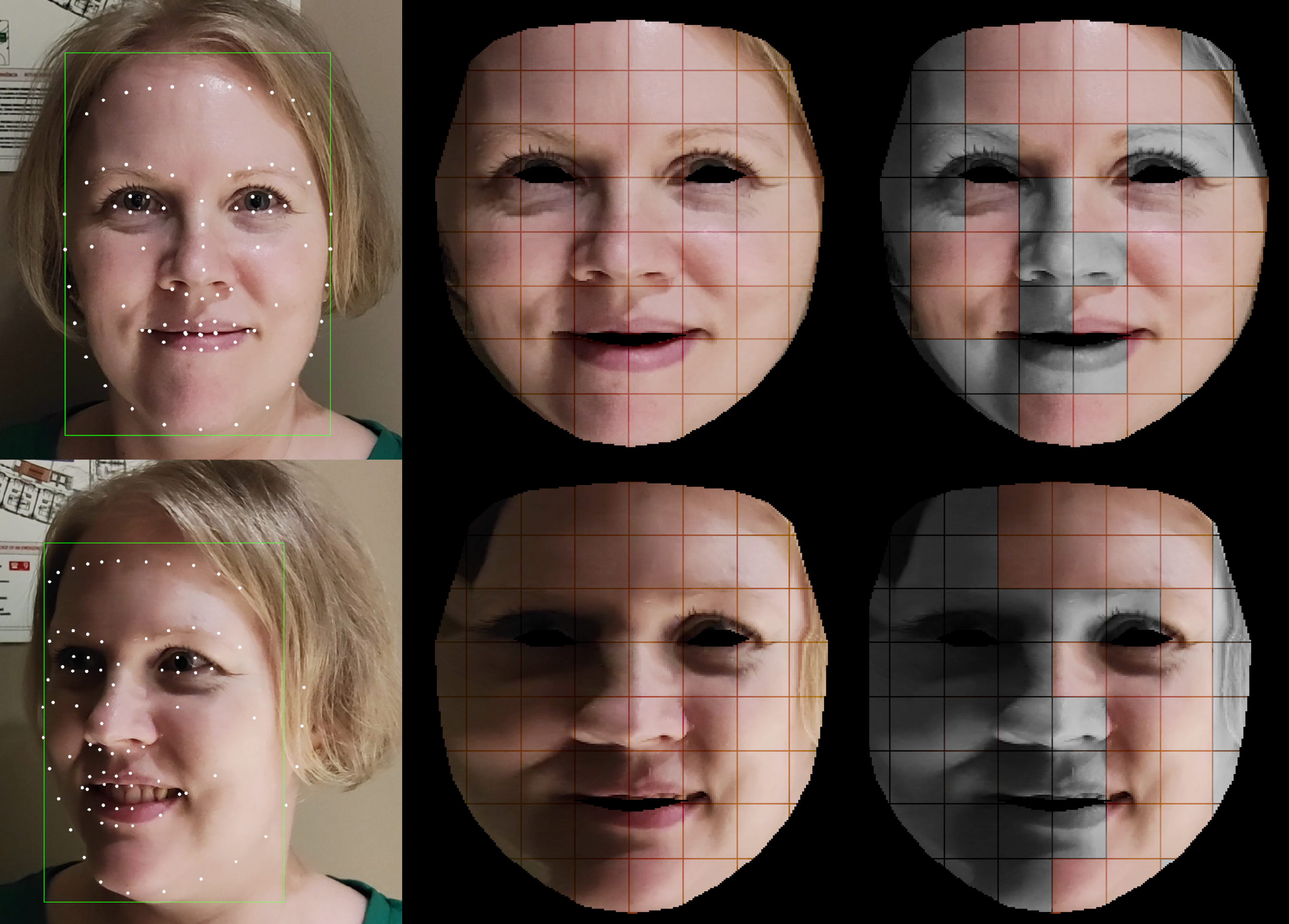}
  \end{center}
  \caption{Comparison of candidate regions of interest in a frontal face under good lighting conditions (top row) and a non-frontal face under suboptimal lighting conditions (bottom row). The system selects 32 regions in the frontal face and 20 regions in the non-frontal face. Grey regions are discarded.}
  \label{fig:GRID_ROIs_Normalized}
\end{figure}

The DMRS process starts just after obtaining a segmented and normalized face in the previous block of the pipeline by dividing the normalized face into a matrix of $n x n$ rectangles (regions of interest) that contains a spatio-temporal representation of the face, as depicted in Figure \ref{fig:GRID_ROIs_Normalized}. Each area in the grid represents a signal in a sequence of frames.

Next, the process continues the analysis by computing several statistical parameters on each candidate region and the global face. These statistical parameters are computed using windows of $t$ seconds, both in time and frequency domains. We extract the mean, standard deviation, variance, signal-to-noise ratio (SNR), Katz Fractal dimension (KFD), number of zero-crossings (Z\textsubscript{c}), sample entropy, detrended fluctuation analysis (DFA), and the energy in terms of local power spectral density (PSD). The dynamic selection is loosely based on fractal analysis, which portrays the scale of the randomness or how unpredictable a stochastic process is \cite{FractalSeries2002}. 
 
The first step is an initial pruning that removes those regions that do not contain valuable information. We check the variance changes of every candidate rectangle along the time $t$ and discard those that have zero variance.

Then, the regions are discarded based on the thresholding of Katz's Fractal Dimension (KFD). KFD computes the fractal dimension (FD) of a signal based on the morphology, measuring the degree of irregularity and the sharpness of the waveform \cite{FractalSeries2002}. KFD gives an index $D_{KFD}$ for characterizing the complexity of the signal. KFD value is computed as shown in the following equation:

\vspace{1mm}
\begin{equation} \label{Formula:KatzEquation}
KFD = \frac{\log_{10}(L/a)}{\log_{10}(d/a)} = \frac{\log_{10}(n)}{\log_{10}(d/L) + \log_{10}(n)}
\end{equation}
\vspace{1mm}

where $L$ is the total length of the PPG time series, $a$ is the average of the Euclidean distances between the successive points of the sample, $d$ is the Euclidean distance between the first point in the series and the point that provides the maximum distance with respect to the first point, and $n$ is $L/a$. We calculate then the relative D\textsubscript{KFD} value by dividing the KFD index of a specific facial region by the global KFD index derived from the entire face, as shown in the following equation:

\vspace{1mm}
\begin{equation} \label{Formula:RelativeKFD}
D_{KFD} = \frac{KFD_{region}}{KFD_{global}}
\end{equation}
\vspace{1mm}

This relative value plays an important role in identifying the regions that contribute meaningful information to the final rPPG signal while addressing the inherent challenges posed by facial regions, such as occlusions, low light conditions, blur, and other factors that could adversely affect signal quality and the complexity of the signal. By selecting regions with a relative D\textsubscript{KFD} value of 0.85 or greater, we ensure the inclusion of regions that exhibit complexity levels comparable to the global PPG signal. This approach is based on the assumption that the primary source of complexity in the global rPPG signal stems from the heart, and regions with similar complexity levels are more likely to provide valuable information for rPPG analysis, despite the presence of factors that could potentially compromise the signal quality. Through this selection process, we effectively discard regions that introduce noise and artifacts in the final rPPG signal. This method provides provide a good balance between discarding low-quality regions and retaining meaningful information.

The analysis continues by using Detrended Fluctuation Analysis (DFA), a statistical method widely used to detect intrinsic self-similarity in non-stationary time series, especially in fractal signals. DFA is a modified root mean square analysis of a random walk, designed to compute long and short-range non-uniform correlations in stochastic processes \cite{DFAPhysiological1995}. The method tells if each region rPPG signal shows the expected correlation with the signal from the global face signal and if they are very noisy or contain artifacts of extrinsic trends \cite{Physionet_MIT_ToolKit2000}. The DFA exponent $\alpha$ is interpreted as an estimation of the Hurst parameter, and it is calculated as the slope of a straight line fit to the log-log graph from the fluctuation function. if $\alpha = 0.5$, the time series is uncorrelated. If $0.5 < \alpha < 1$ then there are positive correlations in the time series. If $\alpha < 0.5$ then the time series is anti-correlated. We discard those regions as uncorrelated and negatively correlated.

Upon completing the analysis of each facial region within a sequence of frames (window of $t$ seconds), we obtain a set of valid regions, with a minimum of 2 and a maximum of 32 regions for the final selection step, which is based on energy content. In cases where more than $r_{max}$ regions pass the previous analysis, our method selects the top $r_{max}$ regions with the highest energy. These regions are expected to exhibit less noisy spectral responses within the relevant frequency range, potentially leading to an improved signal-to-noise ratio \cite{stoica2005spectral}\cite{Biosignals2005SpectralAnalysis}. If the number of regions passing the previous analysis is less than $r_{max}$ but greater than one, all these regions are selected. In contrast, when only one or no regions satisfy the prior criteria, the method reverts to selecting the best $r_{max}$ regions by energy content from the initial set of candidate regions. This strategic approach ensures the derivation of the resulting rPPG signal from an optimal number of regions, thus maximizing signal quality across various scenarios. In the last step, the rPPG signals from the chosen regions are combined by summing them in the time domain, generating the ultimate rPPG signal used for heart rate computation during the subsequent spectral analysis phase.

During the experiments, we have comparatively evaluated \textit{Face2PPG} pipeline both in single-region mode (\textit{Face2PPG-Normalized}) and in multi-region mode (\textit{Face2PPG-Multi}).


\section{Orthogonal Matrix Image Transformation}

The RGB to PPG transformation is a critical step in the field of remote photoplethysmography that allows for the extraction of physiological signals from the color skin variations. However, the process is challenging due to the presence of noise and artifacts in the raw RGB signal. To address these challenges, we introduce a novel, robust, and efficient method called Orthogonal Matrix Image Transformation (OMIT), that we integrated in the previous proposed pipelines.

OMIT is grounded on matrix decomposition techniques and aims to generate an orthogonal matrix with linearly uncorrelated components that represent orthonormal components in the RGB color basis. This allows for the accurate recovery of physiological signals. OMIT employs the reduced (or thin) QR factorization \cite{golub2013matrix} in conjunction with Householder Reflections \cite{HouseholderReflections1958}\cite{shao2021householder} to find linear least-squares solutions in the RGB space. Thin QR factorization offers improved memory efficiency and computational speed compared to full QR factorization, particularly for tall and skinny matrices \cite{golub2013matrix}. Furthermore, the Householder Orthogonalization Algorithm provides better numerical stability, computational efficiency, and conditioning than the Gram-Schmidt process \cite{golub2013matrix}\cite{shao2021householder}. These advantages make the Householder Orthogonalization Algorithm particularly suitable for handling noisy or corrupted data matrices and extracting rPPG signals from raw RGB signals with greater accuracy and efficiency \cite{QR_RecoverCorruptedImages2019}\cite{DenoisingQRD2016}.

The mathematical foundation of OMIT is based on the QR decomposition as shown in Equation \ref{Formula:OMIT}:

\begin{equation} \label{Formula:OMIT}
A = QR
\vspace{3mm}
\end{equation}

where $A \in {\rm I\!R}^{n \times 3}$ represents the input RGB matrix, $Q \in {\rm I\!R}^{n \times 3}$ denotes the orthonormal basis for the column space of A, and $R \in {\rm I\!R}^{3 \times 3}$ is an upper triangular matrix that contains the coefficients to express the columns of A as linear combination of the basis vectors in $Q$. We then use the orthogonal matrix Q to compute a projection matrix that allows us to extract the BVP signal from the input matrix A.

\begin{figure}[h!]
  \begin{center}
    \includegraphics*[width=0.49\textwidth]{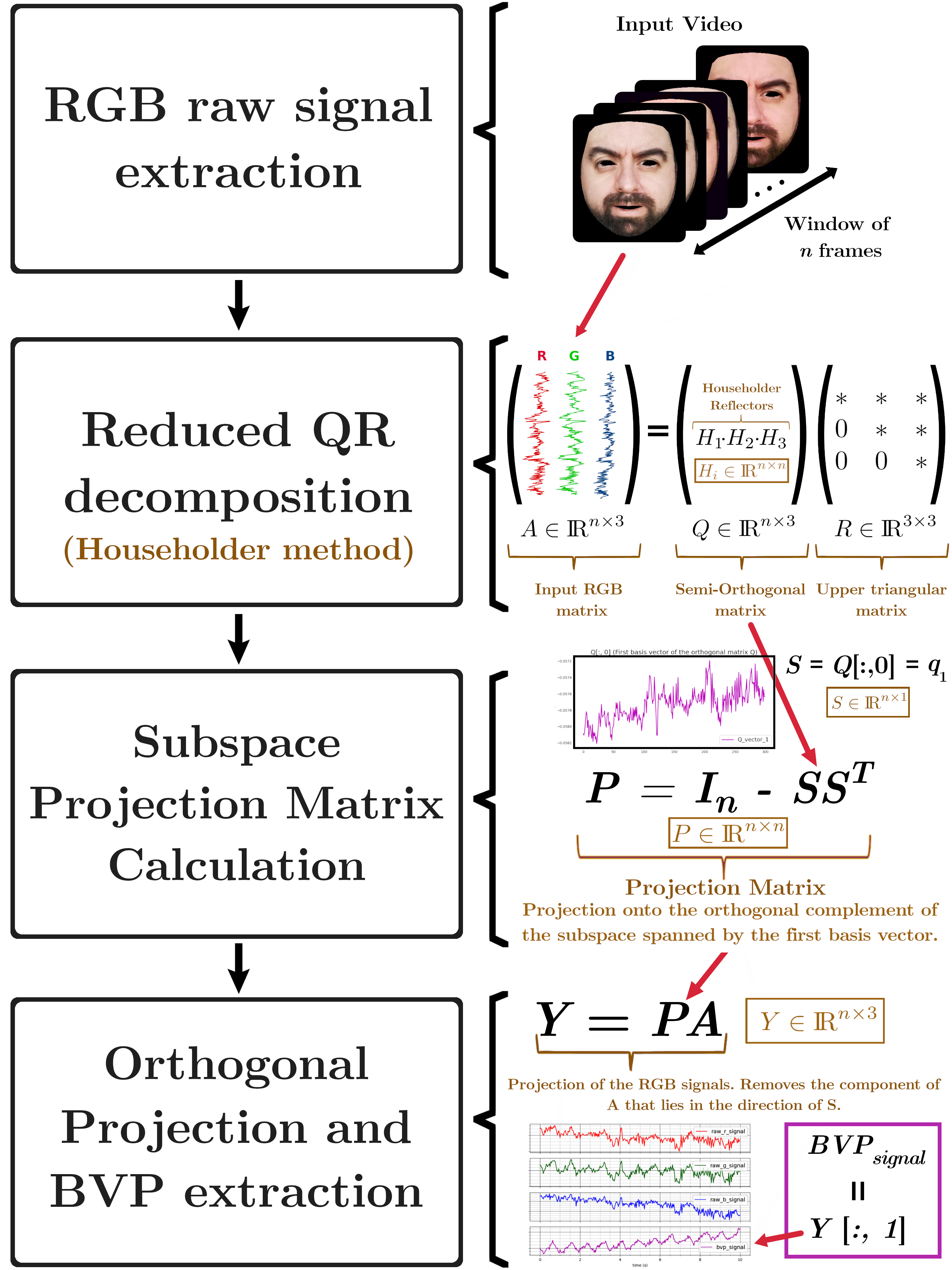}
  \end{center}
  \caption{Process steps to convert a RGB signal to BVP signal using OMIT method.}
  \label{fig:OMITSteps}
\end{figure}

\vspace{3mm}
The OMIT method is composed of the following key steps, illustrated in Figure \ref{fig:OMITSteps}:

\begin{enumerate} 
\item \textbf{Reduced QR decomposition using Householder Reflections}: Compute the thin QR decomposition of the input RGB matrix \cite{golub2013matrix}\cite{QRFactorization1992Application}. For a given input matrix A of dimensions $n \times 3$, the Householder Reflectors ($H_i$) are computed iteratively, transforming A into an upper triangular matrix R. In each iteration, $H_i$ is an $n \times n$ matrix designed to eliminate the elements below the diagonal in the \textit{i-th} column of A or its intermediate form. After $k$ iterations (in our case, $k = 3$), the product of these $H_i$ matrices results in the orthogonal matrix Q, while the transformed A becomes the upper triangular matrix R. Mathematically, this can be expressed as $Q = H_1H_2...H_k$. The Q matrix is semi-orthogonal, meaning that its columns are orthonormal, i.e., they are orthogonal and have unit norm. In our case, Q is an $n \times 3$ matrix, and its columns ($q_1$, $q_2$, $q_3$) form an orthonormal basis for the column space of the input RGB matrix A. The first column, $q_1$, represents the direction in the RGB space that captures the most significant variations in the input data. In the context of rPPG, these variations are typically associated with changes in skin color due to ambient lighting, camera sensor noise, facial movements and other artifacts.
\item \textbf{Subspace projection matrix calculation}: The first column of Q, denoted as S, is used to compute the projection matrix P. This step aims to create an orthogonal subspace to the direction of S, which is computed as $P = I_n - SS^{\mathsmaller T}$. The projection matrix P is an $n\times n$ matrix calculated as the difference between the identity matrix of dimensions \textit{n} ($I_n$) and the outer product of vector S with itself. This matrix is designed to project the input data onto a subspace orthogonal to S. In other words, P removes the contributions associated with the dominant variations in the input data (captured by $q_1$), which are typically unrelated to the BVP signal.
\item \textbf{Orthogonal Projection and BVP extraction}: In the Orthogonal Projection step, the input data (RGB matrix) is projected onto a subspace orthogonal to $q_1$ using the calculated projection matrix P. This process is mathematically represented by the equation $Y = PA$. The purpose of this step is to remove the contributions associated with the dominant variations in the input data, which typically correspond to factors such as lighting conditions and facial movements reflected in the three color channels. The orthogonal projection has significant implications in the OMIT method as it effectively separates the BVP signal from the raw RGB data. The BVP-related information is preserved while the unrelated noise and artifacts are suppressed. The BVP signal is extracted from the second column of the Y matrix instead of the first column, since the first column corresponds to the dominant variations in the input data that were removed during the Orthogonal Projection step. In this context, the second column of the Y matrix represents the processed signal containing the BVP information after removing the dominant variations.
\end{enumerate}

QR decomposition has been extensively applied across various domains such as communications, signal processing, image processing, and machine learning to address challenges associated with corrupted input data matrices due to noise or artifacts \cite{QR_RecoverCorruptedImages2019}\cite{DenoisingQRD2016}.

QR decomposition offers several advantages over other decomposition methods such as Principal Component Analysis (PCA) or Singular Value Decomposition (SVD) \cite{QRFactorization1992Application}\cite{Sharma2013QR}. It is computationally efficient, mathematically stable, and robust to corrupted input data matrices due to noise or artifacts \cite{CHEN2012QRSIMOvsSVD}\cite{DenoisingQRD2016}\cite{QR_RecoverCorruptedImages2019}. Furthermore, it is a widely used method in fields such as communications \cite{Park2021QRMIMO}, signal processing \cite{Barnov2017QRBeamformingIntel}, image processing \cite{Amintoosi2007QRBackgroundSub}, and machine learning \cite{roberts2020qrBackPropagation} and is available in high-performance linear algebra libraries such as LAPACK \cite{LAPACKLib} and Intel\textsuperscript{\tiny\textregistered} MKL \cite{MKLIntel}, making it suitable for real-time applications \cite{Langhammer2018QRFPGAs}. Leveraging these advantages, OMIT produces an orthogonal matrix with linearly uncorrelated components, effectively segregating the rPPG signal from the original RGB data. This approach allows OMIT to enable more accurate and efficient extraction of the concealed blood pulse signal from the raw RGB data, paving the way for enhanced rPPG signal processing and analysis. By incorporating the robustness and stability of QR decomposition, OMIT outperforms EVD and other decomposition techniques, making it a compelling choice for rPPG signal extraction.

\section{Benchmark datasets and Evaluation Metrics}
\label{sec:evaluation}

\textcolor{black}{To evaluate the proposed methodology, we follow an extensive evaluation assessment as presented in the literature \cite{pyVHR2020}. Our evaluation includes interfaces to work on six publicly available datasets:}


\textcolor{black}{\textbf{PURE} is a database that contains 10 subjects performing several controlled head motions \cite{PUREDatabase2014}. The session was recorded using 6 different setups (steady, talking, slow translation, fast translation, slow rotation, and medium rotation), resulting in 60 sequences of 1 minute each. The videos were captured using an industrial-grade (\textit{eco274CVGE} camera by \textit{SVS-Vistek}) at a sampling rate of 30 Hz with an uncompressed cropped resolution of 640x480 pixels and an approximate average distance of 1.1 meters. The reference pulse data was captured in parallel using a contact-based FDA-approved fingertip pulse oximeter (\textit{pulox CMS50E}) with a sampling rate of 60 Hz.}

\textbf{COHFACE} is a remote photoplethysmography (rPPG) dataset that contains RGB videos with faces synchronized with heart rate and breathing rate of the recorded subjects \cite{COHFACE2017}. It contains videos of 40 subjects (12 females and 28 males). The video sequences were recorded using a  (\textit{Logitech HD C525}) webcam at a sampling rate of 20 Hz and a resolution of 640x480 pixels. The database includes a total of 160 videos, of approximately 1 minute. Reference physiological data was recorded using medical-grade equipment.

\begin{figure}[ht!]
  \begin{center}
    \includegraphics*[width=0.48\textwidth]{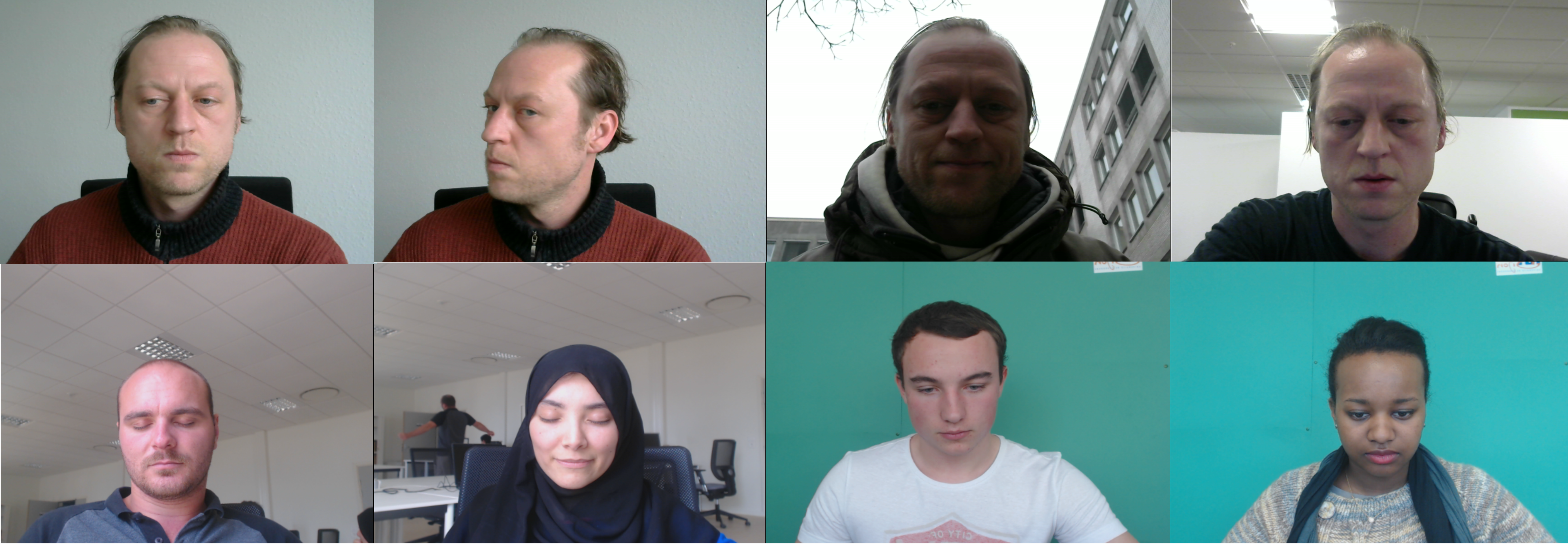}
  \end{center}
  \caption{In the first row, LGI-PPGI database  \cite{LGIMethod2018} contains four different scenario recordings. From left to right: 1) Resting, 2) Rotation or Head Motions, 3) Talking and 4) Gym. In the second row, UBFC-RPPG database \cite{UBFCDatabase2019} contains mostly two different scenarios. UBFC1 contains videos recorded in an office room under unconstrained conditions and natural light (from left to right the first two images). UBFC2 contains videos under controlled conditions but performing a stress task (from left to right, the last two images).}
  \label{fig:UBFC_Scenarios}
\end{figure}

\textbf{LGI-PPGI-Face-Video-Database} is a database that contains 25 subjects, but only 6 were released officially \cite{LGIMethod2018}. It was recorded using a \textit{Logitech HD C270} webcam at a sampling rate of 25 Hz and a resolution of 640x480 pixels, in uncompressed format with auto-exposure. Reference physiological measurements were recorded at the same time using a contact-based FDA-approved fingertip pulse oximeter (\textit{pulox CMS50E}) with a sampling rate of 60 Hz. The database contains subjects in four different scenarios: resting, rotation, talking in the street, and gym. An image with the four scenarios is depicted in Figure \ref{fig:UBFC_Scenarios}.

\textbf{UBFC-RPPG Video dataset} is a rPPG database comprised by two different datasets: UBFC1 and UBFC2 \cite{UBFCDatabase2019}. UBFC1 contains 8 videos where the participants were asked to sit still in an office room under unconstrained conditions and natural light. UBFC2 contains 42 videos under constrained conditions and inducing changes in the BVP by asking participants to perform mathematical games. The database presents a wide variety of ethnicities with different facial skin tones as shown in the Figure \ref{fig:UBFC_Scenarios}.

The database was recorded using a webcam (Logitech C920 HD Pro) at a sampling rate of 30 Hz and a resolution of 640x480 pixels in uncompressed 8-bit RGB format. The duration of each video is approximately two minutes long. The reference physiological data was synchronized and recorded at the same time using a contact-based FDA-approved fingertip pulse oximeter (\textit{pulox CMS50E}) with a sampling rate of 60 Hz.

\textbf{MAHNOB-HCI} is a multimodal database captured mainly for emotion recognition \cite{MAHNOB2021Soleymani}. The database contains 27 young healthy participants, 16 females and 11 males. The database was recorded with several cameras. The frontal camera was an \textit{Allied Vision Stingray F-046C} colour camera with a resolution of 780x540 pixels at 60 frames per second. The videos in MAHNOB-HCI database are highly compressed in H.264/MPEG-4, making them very challenging for the extraction of remote PPG signals. The reference signals were captured using an ECG sensor from the \textit{Biosemi active II} system with active electrodes. The database includes 527 facial videos with corresponding physiological signals. In our evaluation of the different pipelines (Table \ref{tab:PipelinesEvaluationPPG}), we used a smaller subset of 36 videos from the MAHNOB-HCI database, as suggested in \cite{pyVHR2020}, to ensure a direct and fair comparison with the results first presented in \cite{pyVHR2020}. For the comparisons of the state of the art (Table \ref{tab:table_state_of_the_art}, we employed the full dataset of 527 videos to compare our pipeline with other state-of-the-art approaches, providing a comprehensive assessment of our method's performance.

The evaluation of the datasets is done by comparing the estimations of the heart rates of both the extracted rPPG signal and the reference ECG or PPG signal. The evaluation includes both error and statistical analysis.  We use three standard metrics that measure the discrepancy between our predicted heart rate $\hat{h}(t)$ and the reference heart rate $h(t)$. The standard metrics used to compute it are Mean Absolute Error (MAE), Root-Mean-Square Error (RMSE), and Pearson Correlation Coefficient (PCC) of the heart-rate envelope. Our primary objective is to advance unsupervised rPPG extraction techniques in challenging conditions, rather than pursuing waveform similarity with ground truth PPG signals, as is the case with deep learning-based methods. Given the anatomical differences in blood perfusion waveforms between the face and finger \cite{Allen2004PPGsBody}, we employ PCC between heart rate envelopes as a more suitable evaluation metric, instead of direct waveform comparisons.


\subsection{Reference data, ground-truth and evaluation protocol} \label{PPG2HR_Protocols}

The most common source of BVP reference data in the datasets is PPG data from contact-based pulse oximeters. In most of the cases, these data is already filtered and does not require further preprocessing. In order to extract heart rate or other HRV parameters, the reference signals are processed using spectral analysis. 

In the evaluation, we compute the error by comparing the heart rate and HRV parameters extracted from the reference (ground-truth) signals and the recovered PPG signal. Although direct comparison of signals (e.g. morphology) would be also possible, the fundamental differences between the extracted rPPG and the reference signals in terms of delay and scale due to different body measurement points and diverse collection devices make this comparison not very meaningful \cite{Allen2004PPGsBody}. 

\begin{figure}[ht!]
  \begin{center}
    \includegraphics*[width=0.49\textwidth]{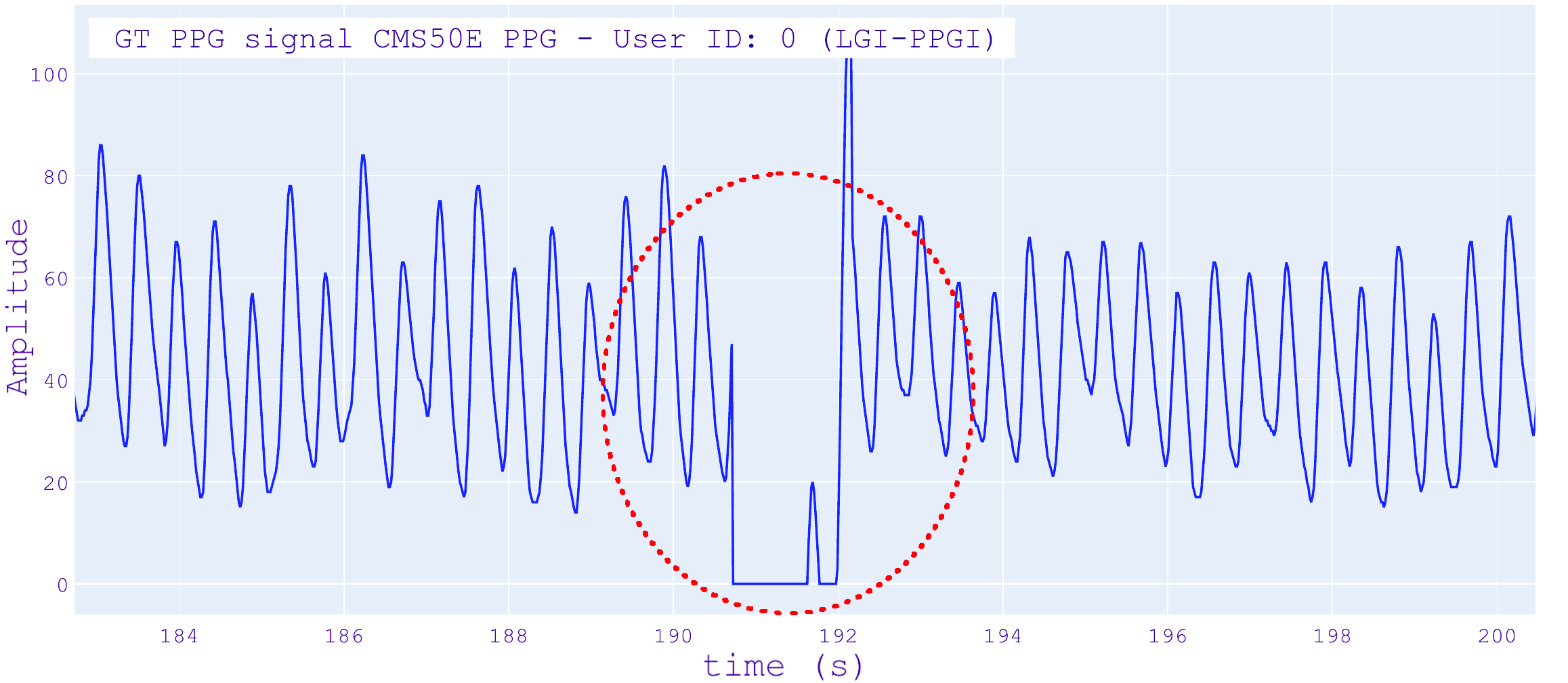}
  \end{center}
  \caption{Section of a reference PPG signal recorded with a fingertip contact-based pulse oximeter where we can observe an error of $\approx$ 2 seconds, probably due to movement of the finger and consequent lost of the signal tracking.}
  \label{fig:GT_Signal_Error}
\end{figure}

We have observed that the reference data offered in the datasets is not completely free of problems. For example, Figure \ref{fig:GT_Signal_Error}, shows an example reference signal with a gap of approximately 2 seconds. These issues, caused by small deficiencies in data collection can lead to unfair disagreements in terms of error, especially for unsupervised methods.

\section{Experimental results}
\label{sec:results}

We evaluate and analyse the proposed methodologies and pipelines to extract remote photoplethysmography signals from all benchmark databases. We compare the results across different improved processing pipelines and compare them with the state of the art for both supervised and unsupervised methods.The experiments are performed using a computer that includes an AMD\textsuperscript{\tiny\textregistered} Ryzen(TM) 3700X 8-core processor at 3.6GHz.

\subsection{Hyperparameters and configuration}
Our framework is based on separate configuration files, in a similar manner as other frameworks \cite{pyVHR2020}. These files contain the parameters that govern the pipeline and its components. In our experiments, we set the values for of each pipeline component as follows: \textbf{Face detection} uses a DNN OpenCV face detector with the default Tensorflow model. \textbf{Face alignment} uses the DAN algorithm with one of the default models provided by the authors (\textit{DAN-Menpo.npz}) \cite{DANKowalskiNT17}. Real-time configurations can use a modified ERT model \cite{AlvarezBordallo2021FaceAlignment}. \textbf{DMRS} uses a grid matrix of $n=9x9$. For D\textsubscript{KFD}, we have selected the signals with a threshold of more than 0.85, and for DFA we set an $\alpha$ threshold between 0.75 and 1.0. We set the maximum number valid of regions $r_{max}=32$. \textbf{Filtering} is performed using FIR filters with Kaiser windows, with the parameter $\beta = 25$. The filters use a bandpasss configuration between 0.75 and 4 Hz (corresponding to 45-240 bpms). \textbf{Signal Windowing} uses sliding windows of 10 seconds and 1 second steps (9 seconds overlap).

\subsection{Quantitative results}
\definecolor{blue}{HTML}{0000ff}
We provide an extensive evaluation of our three proposed pipelines and compare them with those of the baseline \cite{pyVHR2020} with the standard configuration. We obtain results in six datasets. All datasets are comprised by videos with VGA resolution, but in two of them are heavily compressed (MAHNOB and COHFACE). 

We measure the performance by computing the average of the MAE, the standard deviation of the MAE, and the median of the Pearson Correlation Coefficient of the envelope of the heart rate. We evaluate the pipelines using ten rPPG methods (RGB to PPG signal conversion methods), including our proposed OMIT conversion. The results detail the impact of the improvements in each pipeline as shown in Table \ref{tab:PipelinesEvaluationPPG}.

\definecolor{orange}{HTML}{C2570E}
\definecolor{bluet}{HTML}{294483}
\definecolor{green}{HTML}{298323}
\definecolor{red}{HTML}{A31313}
\setlength{\tabcolsep}{0.75em}
\begin{table*}[ht!]
\begin{center}
  \caption{Error comparison between the baseline pipeline and our three proposed improved pipelines.}
  \label{tab:PipelinesEvaluationPPG}
  \scalebox{0.6}{
  \begin{tabular}{clcccccccccccc} 
  \hline\hline\noalign{\smallskip}
    &  \multicolumn{1}{c}{} 
    &  \multicolumn{12}{c}{Databases} \\

\cmidrule(lr){3-14}
       \multicolumn{1}{c}{} 
    &  \multicolumn{1}{c}{} 
    &  \multicolumn{2}{c}{LGI-PPGI} 
    &  \multicolumn{2}{c}{COHFACE} 
    &  \multicolumn{2}{c}{MAHNOB} 
    &  \multicolumn{2}{c}{PURE} 
    &  \multicolumn{2}{c}{UBFC1}
    &  \multicolumn{2}{c}{UBFC2} \\

\cmidrule(lr){3-4}
\cmidrule(lr){5-6}
\cmidrule(lr){7-8}
\cmidrule(lr){9-10}
\cmidrule(lr){11-12}
\cmidrule(lr){13-14}
       \multicolumn{1}{c}{Pipeline} 
    &  \multicolumn{1}{c}{PPG Method}
    &  \multicolumn{1}{c}{MAE ± SD}
    &  \multicolumn{1}{c}{PCC}
     &  \multicolumn{1}{c}{MAE ± SD}
    &  \multicolumn{1}{c}{PCC} 
     &  \multicolumn{1}{c}{MAE ± SD}
    &  \multicolumn{1}{c}{PCC} 
     &  \multicolumn{1}{c}{MAE ± SD}
    &  \multicolumn{1}{c}{PCC} 
     &  \multicolumn{1}{c}{MAE ± SD}
    &  \multicolumn{1}{c}{PCC} 
     &  \multicolumn{1}{c}{MAE ± SD}
    &  \multicolumn{1}{c}{PCC} \\

\noalign{\smallskip}\hline\hline\noalign{\smallskip}
   \multirow{10}{*}{\rotatebox[origin=c]{90}{\makecell{Baseline \\ \small(+ PPG Method)}}} 
        & \textcolor{bluet}{GREEN (2008)}  & 16.3 ± 11.4 & 0.23 & 12.3 ± 7.2 & 0.05 & 15.3 ± 7.5 & 0.01 & 5.8 ± 12.9 & 0.64 & 15.9 ± 10.4 & -0.04 & 8.1 ± 11.7 & 0.60\\
        
        & \textcolor{bluet}{ICA (2010)}  & 11.2 ± 11.4 & 0.22 & 11.1 ± 5.8 & 0.20 & 14.9 ± 6.2 & 0.17 & 4.4 ± 5.2 & 0.36 & 5.8 ± 6.0 & 0.44 & 10.4 ± 7.2 & 0.41 \\
        
        & \textcolor{bluet}{PCA (2011)} & 10.9 ± 12.1 & 0.22 & \textbf{10.4 ± 5.7} & 0.27 & \textbf{11.0 ± 5.4} & 0.17 & 1.9 ± 1.7 & 0.77 & 4.5 ± 1.8 & 0.43 & 5.2 ± 3.9 & 0.58\\

        & \textcolor{bluet}{CHROM (2013)}  & 11.1 ± 14.2 & 0.39 & 12.4 ± 7.0 & 0.03 & 18.6 ± 9.0 & 0.06 & \textbf{1.6 ± 2.0} & 0.83 & 2.2 ± 0.8 & 0.71 & 3.1 ± 2.5 & 0.79 \\
        
        & \textcolor{bluet}{PBV (2014)}  & 17.4 ± 13.1 & 0.03 & 13.2 ± 6.5 & 0.05 & 18.6 ± 6.5 & -0.02 & 3.6 ± 3.7 & 0.45 & 8.9 ± 6.0 & 0.30 & 5.5 ± 7.0 & 0.63 \\
    
        & \textcolor{bluet}{2SR (2016)}  & 10.1 ± 8.9 & 0.31 & 11.8 ± 7.5 & 0.02 & 17.5 ± 5.3 & 0.05 & 3.7 ± 7.1 & 0.73 & 4.1 ± 3.8 & 0.64 & 5.7 ± 5.6 & 0.40 \\
        
        & \textcolor{bluet}{LAB (2016)}  & 10.9 ± 10.8 & 0.35 & 11.2 ± 5.6 & 0.03 & 20.5 ± 10.3 & 0.05 & 2.4 ± 3.5 & 0.64 & 7.5 ± 7.8 & 0.39 & 12.5 ± 13.2 & 0.39\\
        
        & \textcolor{bluet}{POS (2017)} & 11.8 ± 14.4 & 0.38 & 11.9 ± 7.1 & 0.05 & 19.4 ± 5.5 & 0.02 & 1.9 ± 3.1 & 0.85 & \textbf{1.8 ± 0.4} & 0.87 & \textbf{1.9 ± 1.4} & 0.91 \\
        
        & \textcolor{bluet}{LGI (2018)}  & 10.3 ± 13.2 & 0.42 & 12.6 ± 7.1 & -0.01 & 19.0 ± 5.5 & 0.05 & 3.2 ± 3.5 & 0.59 & 2.4 ± 1.0 & 0.70 & 8.7 ± 6.3 & 0.40 \\
        
         & \textcolor{bluet}{OMIT (2022)} & \textbf{9.5 ± 11.1} & 0.48 & 12.4 ± 6.6 & 0.04 & 19.9 ± 9.0 & 0.13 & 2.3 ± 4.9 & 0.65 & 5.9 ± 7.1 & 0.89 & 5.8 ± 8.2 & 0.82 \\

\noalign{\smallskip}\hline\noalign{\smallskip}

\multirow{10}{*}{\rotatebox[origin=c]{90}{\makecell{Improved \\ \small(+ PPG Method)}}} 
        & \textcolor{orange}{GREEN (2008)}  & 9.3 ± 7.9 & 0.27 & 9.6 ± 5.7 & 0.01 & \textbf{12.2 ± 5.8} & 0.16 & 4.8 ± 8.5 & 0.65 & 8.7 ± 10.2 & 0.68 & 5.4 ± 5.9 & 0.81 \\
        
        & \textcolor{orange}{ICA (2010)}  & 7.7 ± 9.3 & 0.30 & 8.8 ± 4.8 & 0.07 & 17.0 ± 7.0 & 0.06 & 1.9 ± 3.9 & 0.86 & 6.3 ± 8.2 & 0.77 & 6.6 ± 8.0 & 0.66 \\
                
        & \textcolor{orange}{PCA (2011)}  & 7.8 ± 9.9 & 0.35 & 8.6 ± 4.8 & 0.04 & 17.2 ± 8.9 & 0.10 & 1.9 ± 4.3 & 0.86 & 5.5 ± 6.5 & 0.59 & 5.0 ± 6.7 & 0.95 \\

        & \textcolor{orange}{CHROM (2013)} & 4.8 ± 5.6 & 0.46 & 8.3 ± 4.4 & 0.03 & 14.4 ± 7.1 & -0.02 & \textbf{1.8 ± 1.9} & 0.83 & \textbf{2.6 ± 2.7} & 0.88 & 5.3 ± 7.6 & 0.94 \\
        
        & \textcolor{orange}{PBV (2014)} & 12.5 ± 9.6 & 0.09 & 9.1 ± 5.8 & 0.10 & 15.8 ± 7.2 & 0.08 & 3.6 ± 8.0 & 0.83 & 7.0 ± 7.9 & 0.65 & 9.4 ± 9.3 & 0.41 \\
    
        & \textcolor{orange}{2SR (2016)}  & 7.5 ± 11.9 & 0.31 & 8.8 ± 6.4 & 0.03 & 12.9 ± 7.0 & 0.10 & 3.5 ± 7.8 & 0.85
        & 3.6 ± 3.8 & 0.83 & 8.6 ± 9.8 & 0.75 \\
        
        & \textcolor{orange}{LAB (2016)}  & 6.0 ± 6.6 & 0.37 & 8.5 ± 4.6 & -0.03 & 16.2 ± 8.3 & -0.01 & 2.6 ± 6.5 & 0.87 & 3.0 ± 2.6 & 0.73 & 7.5 ± 8.8 & 0.39 \\
        
        & \textcolor{orange}{POS (2017)}  & 5.1 ± 6.2 & 0.58 & 7.8 ± 4.6 & 0.04 & 14.8 ± 7.3 & 0.04 & 2.2 ± 9.1 & 0.91 & 2.8 ± 2.7 & 0.89 & \textbf{3.7 ± 7.6} & 0.96 \\
        
        & \textcolor{orange}{LGI (2018)} & 4.7 ± 5.7 & 0.44 & 8.5 ± 3.8 & 0.02 & 14.2 ± 7.2 & -0.09 & 4.2 ± 8.2 & 0.78 & 3.0 ± 2.4 & 0.72 & 4.8 ± 7.0 & 0.90 \\
        
        & \textcolor{orange}{OMIT (2022)} & \textbf{4.4 ± 5.7} & 0.53 & \textbf{7.5 ± 3.9} & 0.03 & 15.3 ± 5.9 & -0.16 & 2.3 ± 9.1 & 0.88 & 3.3 ± 4.0 & 0.93 & 4.3 ± 6.1 & 0.95 \\
                    
\noalign{\smallskip}\hline\noalign{\smallskip}

\multirow{10}{*}{\rotatebox[origin=c]{90}{\makecell{Normalized \\ \small(+ PPG Method)}}} 
       & \textcolor{green}{GREEN (2008)}  & 13.2 ± 9.7 & 0.23 & 13.1 ± 7.5 & -0.03 & 18.7 ± 9.5 & -0.16 & 12.4 ± 13.5 & 0.22 & 7.5 ± 4.4 & 0.46 & 9.0 ± 8.6 & 0.29\\

        & \textcolor{green}{ICA (2010)}  & 10.9 ± 9.2 & 0.33 & 12.3 ± 5.1 & -0.01 & 19.5 ± 6.2 & 0.12 & 3.5 ± 5.7 & 0.73 & 2.1 ± 0.5 & 0.66 & 3.2 ± 5.8 & 0.94\\        
        
        & \textcolor{green}{PCA (2011)}  & 11.9 ± 9.5 & 0.28 & 11.4 ± 5.1 & 0.02 & 19.4 ± 7.3 & 0.06 & 2.9 ± 2.9 & 0.53 & 2.6 ± 1.4 & 0.65 & 2.8 ± 2.9 & 0.90\\
        
        & \textcolor{green}{CHROM (2013)}  & \textbf{6.9 ± 5.8} & 0.47 & 9.5 ± 4.3 & -0.03 & 17.7 ± 9.3 & -0.06 & 1.7 ± 2.1 & 0.78 & 1.5 ± 0.5 & 0.88 & 2.0 ± 2.8 & 0.95 \\
        
        & \textcolor{green}{PBV (2014)}  & 11.1 ± 8.3 & 0.11 & 12.1 ± 5.9 & 0.01 & 18.3 ± 10.1 & 0.03 & 10.3 ± 12.4 & 0.37 & 1.7 ± 0.7 & 0.88 & 4.3 ± 4.4 & 0.85\\
    
        & \textcolor{green}{2SR (2016)}  & 12.0 ± 9.6 & 0.26 & \textbf{9.1 ± 4.6} & 0.03 & 15.2 ± 7.9 & -0.03 & \textbf{1.3 ± 1.7} & 0.87 & 1.7 ± 0.6 & 0.95 & 1.7 ± 1.2 & 0.93\\
        
        & \textcolor{green}{LAB (2016)}  & 8.0 ± 6.9 & 0.44 & 10.1 ± 4.7 & -0.03 & 17.8 ± 11.3 & -0.01 & 5.9 ± 7.5 & 0.49 & 1.9 ± 1.4 & 0.79 & 5.3 ± 5.7 & 0.62\\
        
        & \textcolor{green}{POS (2017)}  & 8.7 ± 8.4 & 0.50 & 9.4 ± 4.8 & 0.10 & 18.1 ± 9.1 & -0.07 & 1.5 ± 2.7 & 0.85 & \textbf{1.2 ± 0.4} & 0.94 & \textbf{1.4 ± 1.5} & 0.95\\
        
        & \textcolor{green}{LGI (2018)} & 8.8 ± 7.3 & 0.42 & 9.2 ± 4.3 & -0.01 & 15.2 ± 7.0 & 0.01 & 2.6 ± 3.5 & 0.63 & 1.8 ± 0.6 & 0.76 & 3.6 ± 5.3 & 0.85\\
        
        & \textcolor{green}{OMIT (2022)} & 9.0 ± 8.0 & 0.39 & 10.6 ± 5.2 & -0.06 & \textbf{14.2 ± 6.3} & -0.08 & 2.2 ± 3.2 & 0.80 & 1.4 ± 0.5 & 0.91 & 3.4 ± 5.3 & 0.87\\

\noalign{\smallskip}\hline\noalign{\smallskip}

\multirow{10}{*}{\rotatebox[origin=c]{90}{\makecell{Multi-region \\ \small(+ PPG Method)}}} 
        & \textcolor{red}{GREEN (2008)}  & 8.3 ± 7.3 & 0.26 & 9.1 ± 6.1 & 0.05 & 12.9 ± 11.0 & 0.21 & 5.4 ± 8.8 & 0.47 & 3.8 ± 2.4 & 0.59 & 1.4 ± 1.7 & 0.94 \\

        & \textcolor{red}{ICA (2010)}  & 9.2 ± 8.1 & 0.16 & 8.6 ± 4.3 & 0.03 & 14.3 ± 8.3 & 0.09 & 2.4 ± 2.6 & 0.71 & 1.5 ± 0.6 & 0.90 & 3.2 ± 3.3 & 0.75\\
        
        & \textcolor{red}{PCA (2011)}  & 9.5 ± 9.7 & 0.22 & 8.6 ± 4.0 & 0.04 & 16.3 ± 8.1 & 0.17 & 2.4 ± 3.1 & 0.72 & 2.0 ± 1.0 & 0.75 & 3.5 ± 4.0 & 0.67 \\
        
        & \textcolor{red}{CHROM (2013)}  & \textcolor{blue}{\textbf{3.9 ± 3.6}} & 0.58 & 8.8 ± 4.5 & 0.03 & 12.6 ± 6.2 & 0.10 & \textcolor{blue}{\textbf{1.2 ± 0.9}} & 0.84 & 0.8 ± 0.5 & 0.98 & 1.5 ± 1.5 & 0.95\\
        
        & \textcolor{red}{PBV (2014)}  & 9.5 ± 7.6 & 0.24 & 9.5 ± 4.6 & 0.09 & 15.9 ± 7.7 & 0.06 & 4.9 ± 7.4 & 0.53 & 3.7 ± 2.9 & 0.58 & 4.7 ± 5.2 & 0.65\\
    
        & \textcolor{red}{2SR (2016)}  & 14.8 ± 8.6 & 0.08 & 9.1 ± 4.5 & 0.09 & 18.0 ± 8.7 & 0.06 & 5.7 ± 7.7 & 0.49 & 1.3 ± 0.5 & 0.91 & 6.6 ± 7.7 & 0.46\\
        
        & \textcolor{red}{LAB (2016)}  & 6.4 ± 5.9 & 0.35 & 9.5 ± 4.4 & -0.03 & 14.6 ± 7.1 & 0.03 & 3.0 ± 4.6 & 0.68 & 1.0 ± 0.5 & 0.96 & 2.5 ± 2.4 & 0.81 \\
        
        & \textcolor{red}{POS (2017)}  & 4.5 ± 3.3 & 0.57 & 8.0 ± 4.4 & 0.06 & 13.6 ± 6.7 & 0.05 & 1.4 ± 1.6 & 0.86 & 0.9 ± 0.4 & 0.96 & \textcolor{blue}{\textbf{0.9 ± 0.9}} & 0.98\\
        
        & \textcolor{red}{LGI (2018)}  & 4.5 ± 3.1 & 0.56 & \textcolor{blue}{\textbf{7.5 ± 3.5}} & 0.11 & 10.8 ± 6.0 & 0.07 & 1.7 ± 1.9 & 0.72 & 1.4 ± 0.9 & 0.92 & 1.5 ± 1.6 & 0.96\\
        
        & \textcolor{red}{OMIT (2022)}  & 4.4 ± 3.1 & 0.62 & 8.0 ± 4.1 & 0.08 & \textcolor{blue}{\textbf{9.3 ± 4.1}} & 0.08 & 1.7 ± 2.6 & 0.86 & \textcolor{blue}{\textbf{0.8 ± 0.4}} & 0.97 & 1.1 ± 1.2 & 0.98 \\

\noalign{\smallskip}\hline\hline
\end{tabular}}
\end{center}
\vspace{1mm}
\footnotesize{The results for MAHNNOB and UBFC2 are computed in a smaller subset of videos, according to \cite{pyVHR2020}.The UBFC2 subset is comprised by 26 videos (Numbers 1, 3, 10, 11, 12, 13, 14, 15, 16, 17, 18, 20, 22, 23, 24, 25, 26, 27, 30, 31, 32, 33, 34, 35, 36 and 37). The MAHNOB subset is composed by 36 videos (Numbers 10, 1198, 1576, 2094, 2226, 2604, 2634, 2888, 3388, 3644, 556, 808, 1042, 1322, 1590, 2114, 2246, 2606, 274, 3006, 3396, 3662, 800, 814, 1184, 152, 1698, 2120, 2346, 2624, 276, 3136, 34, 408, 806 and 926). For LGI-PPGI, COHFACE, UBFC1 and PURE, we use the full set of videos.}
\vspace{-3mm}
\end{table*}

The Multi-region pipeline, with our proposed improvements, achieves the best results across all six datasets, improving MAE, error standard deviation, and Pearson Correlation Coefficient of the heart rate envelope.

Analyzing different rPPG conversion methods, CHROM and POS perform best in uncompressed databases across all pipelines, with OMIT closely following. OMIT works well in highly compressed pipelines, obtaining the best results for the challenging MAHNOB dataset.

Comparing results across datasets, the mean average error varies significantly depending on the data nature. Good quality and static datasets like UBFC or PURE have an error below 2 bpms, with minor differences across videos. For the LGI-PPGI dataset with natural movement, the best average error reaches nearly 4 bpms, with a reasonably high standard deviation. Worst results correspond to lower resolution datasets like MAHNOB and COHFACE, with average errors between 8 and 12 bpms. Heavy video compression and low illumination can cause low SNR and loss of signal subtleties \cite{EffectsVideoCompression_HR_PPG}.

\begin{figure}[ht!]
  \begin{center}
    \includegraphics*[width=0.49\textwidth]{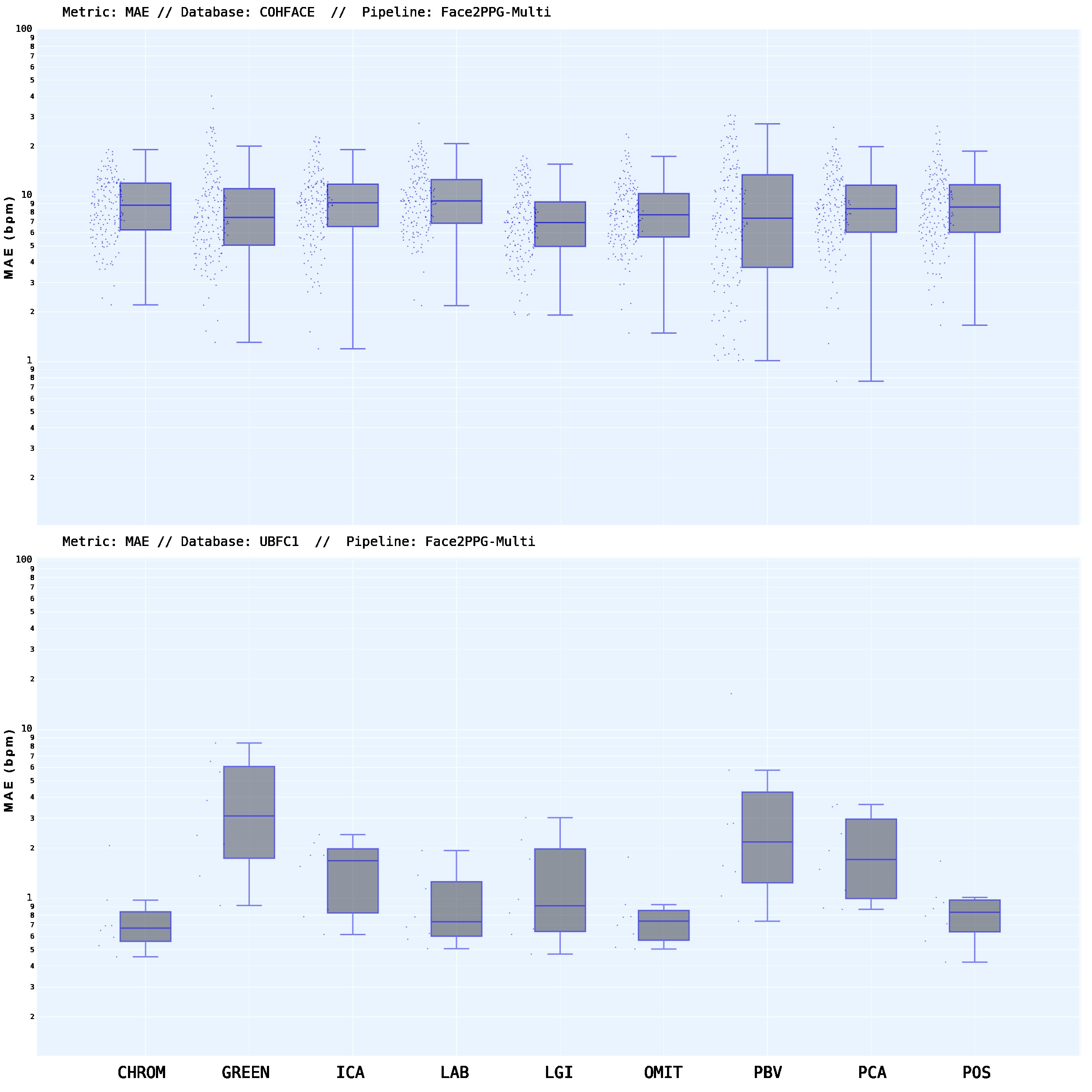}
  \end{center}
  \caption{MAE in logarithmic scale for COHFACE (up) and UBFC1 (down) databases using the Multi-region pipeline with 9 different rPPG methods. The middle line in the boxes indicates the median at every method.}
  \label{fig:MAE_in_COHFACE_Methods}
\end{figure}

In each dataset, the error presents variations along the different videos. This is shown in the standard deviation of the MAE, which reflects this variations. As a more detailed example, we computed the error and the Pearson Correlation Coefficient with its mean, maximum, minimum, and standard deviation for 9 RGB to PPG conversion methods, for the COHFACE and UBFC1 datasets and the Multi-region pipeline. We depict it graphically in Figure \ref{fig:MAE_in_COHFACE_Methods}.

It can be seen that for the high video quality dataset, there is some variability across rPPG conversion methods, while the variability among videos is relatively low. For the heavily compressed dataset, the error and its variability across videos is generally higher, but stable across different rPPG methods. Although some variations across methods, dataset, and pipelines exist, it is possible to conclude that the modification introduced in the pipelines shows a consistent improvement, although those databases with faces mostly still in front of the camera (PURE and UBFC), show only modest improvements when compared with those achieved in complicated datasets (LGI-PPGI, COHFACE, MAHNOB). 


In this context, we argue that the proposed multi-region pipeline shows better performance when presented with videos collected in a more natural environment, especially when they show varied facial expressions, head movements, or illumination changes. To illustrate this, we disaggregate the results obtained on the LGI-PPGI dataset. This dataset is divided in four different scenarios. The resting scenario is a reference scenario where the participants are still in front of the camera with no head or facial movements and illumination, mostly static. In the rotation scenario, the subjects are also in the same setup, but the subjects perform a series of head motions and rotations. In the talking scenario, the subjects are in the wild (mainly in the street), under unconstrained conditions talking in video conference mode with sudden face and head motions. The illumination is natural light with strong back-light conditions in some of the videos, provoking low dynamic range (LDR) images. The last session represents a sports scenario recorded in a gym, where the subjects freely perform a physical exercise on a static bicycle. The illumination is mainly ceiling lights. In some of the videos, we can appreciate the flickering effect due to the ceiling lights.  These last two could represent typical scenarios where we would need to remotely extract physiological signals, such as during video conferences for remote healthcare or sport performance monitoring. We show the results in Table \ref{tab:PerformanceByActivities}. We compare the baseline and multiregion pipelines using our proposed OMIT method for RGB to PPG conversion.

\definecolor{orange}{HTML}{C2570E}
\definecolor{blue}{HTML}{294483}
\definecolor{green}{HTML}{298323}
\definecolor{red}{HTML}{A31313}
\setlength{\tabcolsep}{0.5em}
\begin{table}[ht!]
\begin{center}

\caption{Performance of the Baseline and Multi-region pipelines (LGI-PPGI dataset) for different human activities.}
  \label{tab:PerformanceByActivities}
  \scalebox{0.8}{
  \begin{tabular}{cccccccc} 
  \hline\hline\noalign{\smallskip}

       \multicolumn{1}{c}{} 
    &  \multicolumn{3}{c}{Baseline} 
    &  \multicolumn{3}{c}{Multi-region}  \\

\cmidrule(lr){2-4}
\cmidrule(lr){5-7}

       \multicolumn{1}{c}{Scenario} 
    &  \multicolumn{1}{c}{MAE ± SD}
    &  \multicolumn{1}{c}{PCC}
     &  \multicolumn{1}{c}{RMSE}
    &  \multicolumn{1}{c}{MAE ± SD}
    &  \multicolumn{1}{c}{PCC}
    &  \multicolumn{1}{c}{RMSE} \\

\noalign{\smallskip}\hline\hline\noalign{\smallskip}
        \makecell{Resting}
        & 1.8 ± 1.9 & 0.83 & 2.8 & 1.3 ± 1.3 & 0.87 & 1.7   \\
        \hline\noalign{\smallskip}
        
        \makecell{Rotation}
        & 5.2 ± 3.2 & 0.10 & 9.4 & 5.1 ± 2.8 & 0.58 & 7.6   \\
        \hline\noalign{\smallskip}
        
         \makecell{Talking}
        & 9.5 ± 9.8 & 0.59 & 11.3 & 5.8 ± 3.9 & 0.57 & 7.5   \\
        \hline\noalign{\smallskip}
        
        \makecell{Gym}
        & 23.2 ± 11.4 & 0.16 & 27.6 & 7.5 ± 4.8 & 0.76 & 11.1   \\

\noalign{\smallskip}\hline\hline
\end{tabular}}

\end{center}
\end{table}


\definecolor{blue}{HTML}{0000ff}

\subsection{Qualitative Results}

\begin{figure}[ht]
\begin{center}
\includegraphics*[width=0.49\textwidth]{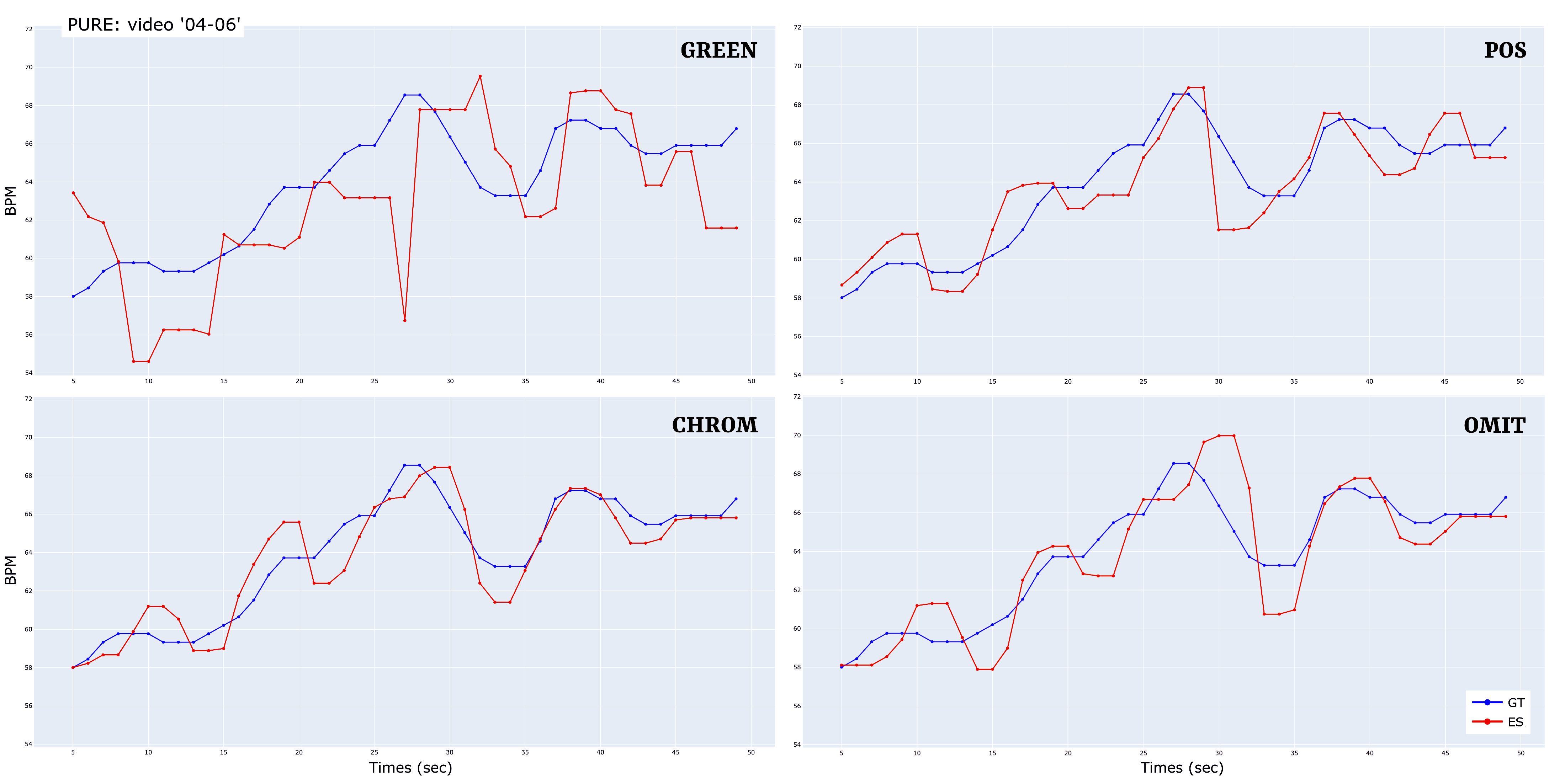}
\end{center}

\caption{Comparison of HR estimation using four rPPG methods with the Multi-region pipeline. Estimated heart rate (red) is extracted from the face, and the contact-based reference PPG signal (blue) is computed on a single PURE dataset video.}
\label{fig:HR_Estimation_Comparison_Methods}

\end{figure}

We provide a visual representation of heart rate estimations from various pipelines and methods in Figure \ref{fig:HR_Estimation_Comparison_Methods}. It shows the performance of four rPPG methods (CHROM, OMIT, POS, and GREEN) using the Multi-region pipeline on a PURE dataset video. CHROM, OMIT, and POS have similar performance, while GREEN struggles to track pulse rate during certain segments. The PCC metric reflects the similarity between the estimated (red) and ground-truth (blue) envelopes.

\begin{figure}[ht]
\vspace{-5mm}
\begin{center}
\includegraphics*[width=0.49\textwidth]{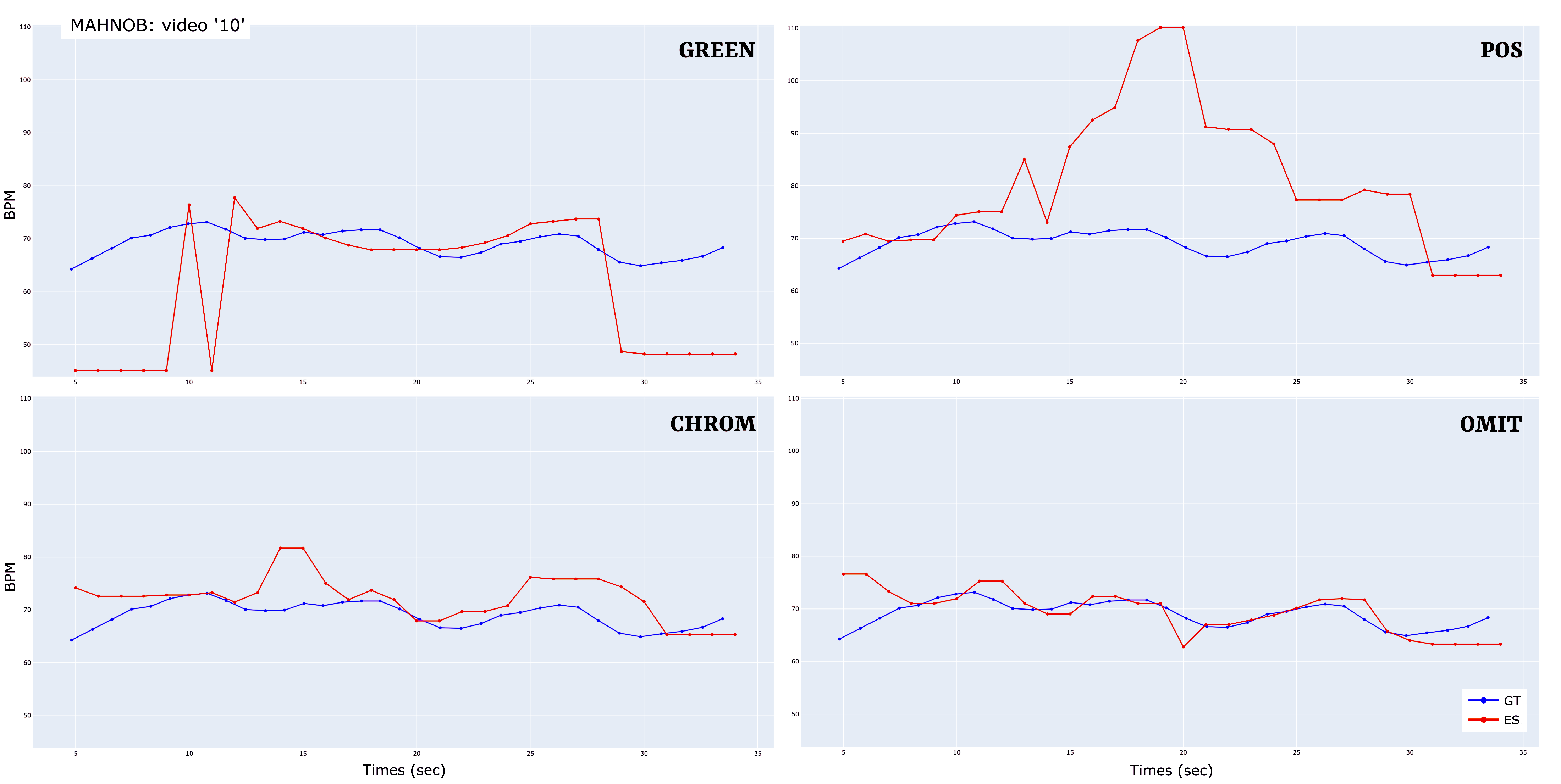}
\end{center}
\caption{Comparison of HR estimation using four rPPG methods with the Multi-region pipeline. Estimated heart rate (red) is extracted from the face, and the contact-based reference PPG signal (blue) is computed on a single MAHNOB video.}
\label{fig:HR_Estimation_Comparison_Methods2}
\vspace{-3mm}
\end{figure}

Figure \ref{fig:HR_Estimation_Comparison_Methods2} shows the performance of four rPPG methods using the Multi-region pipeline on a MAHNOB database video. Despite the subject being static, high compression causes a loss of detail in raw RGB signals. CHROM and OMIT handle compression challenges better than GREEN and POS.

\subsection{Evaluation of the number of regions}

\begin{figure}[ht!]
 \begin{center}
   \includegraphics*[width=0.49\textwidth]{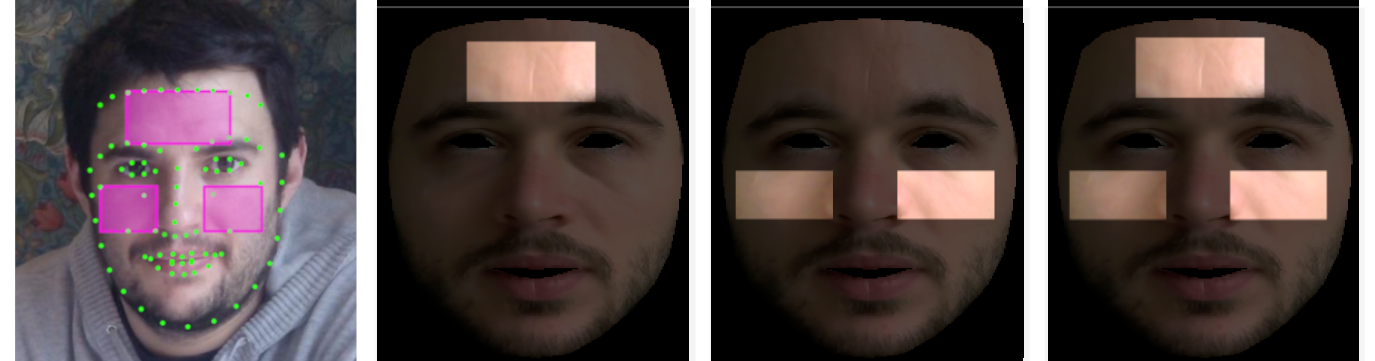}
 \end{center}
 \caption{Region selection based on normalized fixed patches. From left to right: face landmark detection, forehead patch, cheek patches and both combined.}
 \label{fig:Fixed_Patches_Face}
\end{figure}

To evaluate the impact of the DMRS module in the Multi-region pipeline, we have designed a complementary experiment that measures how the results are affected  depending on the number of facial regions used in the initial grid. We use CHROM as the RGB to rPPG conversion method, while all parameters remain the same except the number of initial available regions to select. In the comparison, in addition to region grids, we also include the typical fixed regions of the face, such as the forehead and cheeks, as depicted in Figure \ref{fig:Fixed_Patches_Face}.

For this experiment, the results are depicted in Table \ref{tab:DMRA_Evaluation}. They show how generally, a moderately large number of regions results in smaller errors. The approach using fixed patches shows comparable results to configurations with low number of regions and proved to be still useful in some cases.

\setlength{\tabcolsep}{0.5em}
\begin{table}[ht!]
\begin{center}
  \caption{Impact of the number of regions in rPPG extraction using
  \textit{Face2PPG-Multi} pipeline and CHROM.}
  \label{tab:DMRA_Evaluation}
  \scalebox{0.75}{
  \begin{tabular}{ccccccc} 
  \hline\hline\noalign{\smallskip}

       \multicolumn{1}{c}{} 
    &  \multicolumn{3}{c}{LGI-PPGI} 
    &  \multicolumn{3}{c}{COHFACE}  \\

\cmidrule(lr){2-4}
\cmidrule(lr){5-7}

       \multicolumn{1}{c}{Regions} 
    &  \multicolumn{1}{c}{MAE ± SD}
    &  \multicolumn{1}{c}{PCC}
     &  \multicolumn{1}{c}{RMSE}
    &  \multicolumn{1}{c}{MAE ± SD}
    &  \multicolumn{1}{c}{PCC}
    &  \multicolumn{1}{c}{RMSE} \\

\noalign{\smallskip}\hline\hline\noalign{\smallskip}
        \makecell{6 x 6}
        & 4.3 ± 4.3 & 0.60 & 5.9 & 8.9 ± 4.6 & -0.03 & 11.0   \\
        \hline\noalign{\smallskip}
        
        \makecell{7 x 7}
        & 4.1 ± 3.5 & 0.62 & 5.6 & 8.9 ± 4.4 & 0.01 & 11.0   \\
        \hline\noalign{\smallskip}
        
         \makecell{8 x 8}
        & 4.0 ± 3.4 & 0.64 & 5.7 & 9.0 ± 4.3 & 0.03 & 11.1   \\
        \hline\noalign{\smallskip}
        
         \makecell{9 x 9}
        & 3.9 ± 3.6 & 0.59 & 5.5 & 8.8 ± 4.5 & 0.03 & 11.0   \\
        
        \hline\noalign{\smallskip}
        
        \makecell{10 x 10}
        & 4.0 ± 3.4 & 0.63 & 5.7 & 9.2 ± 4.4 & -0.01 & 11.3   \\
        
        \hline\noalign{\smallskip}
        
        \makecell{11 x 11}
        & 4.2 ± 4.0 & 0.61 & 5.8 & 9.3 ± 4.5 & 0.02 & 11.5   \\
        
       \hline\noalign{\smallskip}
        
        \makecell{Forehead}
        & 5.4 ± 6.3 & 0.63 & 7.5 & 11.6 ± 5.5 & -0.01 & 14.3   \\
        \hline\noalign{\smallskip}
        
        \makecell{Cheeks}
        & 6.5 ± 5.5 & 0.42 & 8.7 & 10.9 ± 4.0 & 0.05 & 13.4   \\
        \hline\noalign{\smallskip}
        
         \makecell{Combined}
        & 5.1 ± 4.5 & 0.60 & 7.3 & 10.1 ± 4.5 & -0.01 & 12.6   \\

\noalign{\smallskip}\hline\hline
\end{tabular}}

\end{center}
\end{table}

\subsection{Comparison with the state of the art}
We compare our proposed pipeline with other supervised learning-based and unsupervised non learning-based methods as presented in the state of the art. We compare the results in terms of both MAE and RMSE and show them in Table  \ref{tab:table_state_of_the_art}. Table \ref{tab:table_state_of_the_art}, shows two sets of results. Face2PPG-Multi\textsubscript{best} results represent the best performance achieved using our multi-region pipeline with various RGB to PPG transformation methods (CHROM, POS, LGI, or OMIT) showing the flexibility of our pipeline. Face2PPG-Multi\textsubscript{omit} results display the outcomes applying just the OMIT method, to highlight its generalizability across diverse datasets.

\definecolor{orange}{HTML}{C2570E}
\definecolor{blue}{HTML}{294483}
\definecolor{green}{HTML}{298323}
\setlength{\tabcolsep}{1.1em}
\begin{table*}[ht!]
\begin{center}
  \caption{Comparison of \textit{Face2PPG-Multi} pipeline with state of the art supervised learning-based (orange) and unsupervised non-learning-based (blue) methods.}
  \label{tab:table_state_of_the_art}
  \scalebox{0.7}{
  \begin{tabular}{clcccccccc} 
  \hline\hline\noalign{\smallskip}
    &  \multicolumn{1}{c}{} 
    &  \multicolumn{6}{c}{Databases} \\

\cmidrule(lr){3-8}
       \multicolumn{1}{c}{Metric} 
    &  \multicolumn{1}{c}{Method} 
    &  \multicolumn{1}{c}{LGI-PPGI} 
    &  \multicolumn{1}{c}{COHFACE} 
    &  \multicolumn{1}{c}{MAHNOB (*)} 
    &  \multicolumn{1}{c}{PURE} 
    &  \multicolumn{1}{c}{UBFC1}
    &  \multicolumn{1}{c}{UBFC2 (*)} \\

\noalign{\smallskip}\hline\hline\noalign{\smallskip}
   \multirow{13}{*}{\rotatebox[origin=c]{90}{\makecell{MAE \\ \small(bpm ± SD)}}} 
        & \textcolor{blue}{\textbf{Face2PPG\textsubscript{best} (Ours)}}  &  \textbf{3.9 ± 3.6} & 7.5 ± 3.5 & 9.3 ± 4.1 & 1.2 ± 0.9 & \textbf{0.8 ± 0.4} & \textbf{0.9 ± 0.9} \\

        & \textcolor{blue}{\textbf{Face2PPG\textsubscript{omit} (Ours)}}  &  4.4 ± 3.1 & 8.0 ± 4.1 & 9.3 ± 4.1 & 1.7 ± 2.6 & \textbf{0.8 ± 0.4} & 1.1 ± 1.2 \\
        
        & \textcolor{blue}{CHROM (2013)} \cite{CHROMMethod2013}  &  9.4 ± 12.6 & 12.4 ± 7.0 & 18.6 ± 9.0 & 1.6 ± 2.0 & 2.2 ± 0.8 & 8.3 ± 10.9  \\ 

        & \textcolor{blue}{Li et al. (2014)} \cite{LiMethod2014} &  - & 19.9 ± n/a & 7.4 ± n/a$^1$ & 28.2 ± n/a & - & - \\

        & \textcolor{blue}{POS (2017)} \cite{POSMethod2017} &  9.7 ± 11.9 & 11.9 ± 7.1 & 19.4 ± 5.5 & 1.9 ± 3.1 & 1.8 ± 0.4 & 7.9 ± 10.4  \\ 
        
        & \textcolor{blue}{FaceRPPG (2020)} \cite{FaceRPPG2020Gudi} &  -  & 14.2 ± 11.6 & 18.6 ± 13.5 & 0.4 ± 0.4$^2$ & - & 2.7 ± 6.7$^2$  \\

        & \textcolor{orange}{HR-CNN (2018)} \cite{HRCNNMethod2018} &  - & 8.1 ± 9.5 & 7.3 ± n/a & 1.8 ± n/a & - & 4.9 ± 5.1  \\
        
        & \textcolor{orange}{DeepPhys (2018)} \cite{DeepPhysMethod2018} & -  & - & 4.6 ± n/a & \textbf{0.8 ± n/a} & - & 6.3 ± 8.7 \\
        
        & \textcolor{orange}{rPPGNet (2019)} \cite{ZitongMethod2019} &  - & - & 5.5 ± n/a & - & - & -  \\
        
        & \textcolor{orange}{Meta-rPPG (2020)} \cite{MetaRPPG2020} &  - & 9.3 ± 11.5 & \textbf{3.0 ± 4.9} & - & - & 6.0 ± 7.1  \\
        
        & \textcolor{orange}{AutoHR (2020)} \cite{AutoHR2020Zitong} &  - & - & 3.8 ± n/a & - & - & -   \\
        
        & \textcolor{orange}{PulseGAN (2021)} \cite{PulseGAN2021} &  - & - & 4.2 ± n/a & 2.3 ± n/a & - & 2.1 ± n/a  \\
        
        & \textcolor{orange}{AND-rPPG (2021)} \cite{ANDrPPG2022} &  - & \textbf{6.8 ± 7.8} & - & - & - & 3.2 ± 4.0  \\

\noalign{\smallskip}\hline\noalign{\smallskip}

\multirow{13}{*}{\rotatebox[origin=c]{90}{\makecell{RMSE \\ \small(bpm ± SD)}}} 
        & \textcolor{blue}{\textbf{Face2PPG\textsubscript{best} (Ours)}} &  \textbf{5.5 ± 5.0} & 9.8 ± 4.6 & 12.5 ± 5.8 & 1.8 ± 1.9 & \textbf{1.1 ± 0.5} & \textbf{2.8 ± 3.0}  \\

        & \textcolor{blue}{\textbf{Face2PPG\textsubscript{omit} (Ours)}} &  6.4 ± 4.9 & 10.5 ± 4.4 & 12.5 ± 5.8 & 2.9 ± 4.6 & \textbf{1.1 ± 0.5} & 3.1 ± 3.8  \\
        
        & \textcolor{blue}{CHROM (2013)} \cite{CHROMMethod2013} &  11.4 ± 13.7 & 16.4 ± 7.9 & 22.1 ± 8.7 & 3.0 ± 4.8 & 4.8 ± 3.4 & 10.3 ± 13.5  \\ 
        
        & \textcolor{blue}{Li et al. (2014)} \cite{LiMethod2014} &  - & 25.6 ± n/a & 7.6 ± 6.9$^1$ & 31.0 ± n/a & - & -  \\

        & \textcolor{blue}{POS (2017)} \cite{POSMethod2017} &  11.9 ± 13.0 & 16.1 ± 8.1 & 27.7 ± 6.2 & 3.6 ± 6.3 & 2.6 ± 0.9 & 9.9 ± 11.7  \\ 
        
        & \textcolor{blue}{FaceRPPG (2020)} \cite{FaceRPPG2020Gudi} &  -  & - & - & - & - & - \\

        & \textcolor{orange}{HR-CNN (2018)} \cite{HRCNNMethod2018}  & - & 10.8 ± n/a & 9.2 ± n/a & 2.4 ± n/a & - & 5.9 ± n/a   \\
        
        & \textcolor{orange}{DeepPhys (2018)} \cite{DeepPhysMethod2018}  & -  & - & - & \textbf{1.5 ± n/a} & - & 10.5 ± n/a  \\
        
        & \textcolor{orange}{rPPGNet (2019)} \cite{ZitongMethod2019}  & - & - & 7.8 ± 7.8 & - & - & - \\ 
        
        & \textcolor{orange}{Meta-rPPG (2020)} \cite{MetaRPPG2020} &  - & 12.3 ± n/a & \textbf{3.7 ± n/a} & - & - & 7.4 ± n/a  \\
        
        & \textcolor{orange}{AutoHR (2020)} \cite{AutoHR2020Zitong} &  - & - & 5.1 ± 4.3 & - & - & -   \\

        & \textcolor{orange}{PulseGAN (2021)} \cite{PulseGAN2021} &  - & - & 6.5 ± n/a & 4.3 ± n/a & - & 4.4 ± n/a   \\
        
        & \textcolor{orange}{AND-rPPG (2021)} \cite{ANDrPPG2022} &  - & \textbf{8.1 ± n/a} & - & - & - & 4.8 ± n/a   \\

\noalign{\smallskip}\hline\hline

\end{tabular}}
\end{center}
\footnotesize{(*) We use the full set of each database for comparison, including MAHNOB (527 videos) and UBFC2 (42 videos).
$^1$ Li et al.'s \cite{LiMethod2014} results were obtained using a smaller MAHNOB subset, removing low-quality PPG sections. Our fair configuration reproduction yields a MAE of 15.5.
$^2$ Gudi et al.'s \cite{FaceRPPG2020Gudi} results use adaptive post-processing filtering. Their manuscript indicates an expected MAE of 1.2 for PURE and 4.1 for UBFC2 when using 10s. Some of the results for deep learning-based methods are referenced from \cite{ANDrPPG2022}\cite{gupta2023radiant}.}
\end{table*}

It can be seen that our method, relying on  the multi-region pipeline obtains better results that all unsupervised non-learning-based methods across all six benchmark datasets. Our results are also comparable to some recent supervised methods that require training on videos taken in similar conditions. Additionally, as shown in paper \cite{Alvarez2023rPPG2Depression}, our method is highly efficient, taking only 17 ms per frame, and under 33 ms with face detection and alignment. This outperforms deep learning methods, which generally require longer processing times per frame, underscoring our approach's computational advantage.


\section{Conclusion}
\label{sec:con}
In this article, we proposed a new unsupervised pipeline for the extraction of BVP signals from facial videos (rPPG). To enable a fair comparative evaluation among methods, we solved a set of smaller technical challenges such as problems with signal synchronization, use of different spectral analysis methods in extracted and reference signals, or inconsistent use of pipeline modules such as face detection and tracking or filtering. We proposed three novel contributions that improve the extraction of rPPG signals, especially in challenging conditions. First, we included a face normalization module, based on facial landmarks and a fixed triangle mesh that allowed the extraction of signals from exactly the same facial regions in a consistent manner. Second, we added the dynamic selection of facial regions that allowed to statistically discard those regions showing noise and artifacts. Finally, we proposed a novel RGB to PPG conversion method that increased the robustness of the extraction against compression artifacts. Our enhanced pipeline works in a purely unsupervised manner, and it is directly applicable in datasets collected in multiple conditions without any need of training data. The proposed pipeline achieves state-of the-art results across multiple databases when compared with other unsupervised methods and shows comparable results to other unsupervised methods.

\section{Acknowledgments}
This research has been supported by the Academy of Finland 6G Flagship program under Grant 346208 and PROFI5 HiDyn under Grant 326291.


\bibliography{references}


\begin{thebibliography}{79}
\ifx \bisbn   \undefined \def \bisbn  #1{ISBN #1}\fi
\ifx \binits  \undefined \def \binits#1{#1}\fi
\ifx \bauthor  \undefined \def \bauthor#1{#1}\fi
\ifx \batitle  \undefined \def \batitle#1{#1}\fi
\ifx \bjtitle  \undefined \def \bjtitle#1{#1}\fi
\ifx \bvolume  \undefined \def \bvolume#1{\textbf{#1}}\fi
\ifx \byear  \undefined \def \byear#1{#1}\fi
\ifx \bissue  \undefined \def \bissue#1{#1}\fi
\ifx \bfpage  \undefined \def \bfpage#1{#1}\fi
\ifx \blpage  \undefined \def \blpage #1{#1}\fi
\ifx \burl  \undefined \def \burl#1{\textsf{#1}}\fi
\ifx \doiurl  \undefined \def \doiurl#1{\url{https://doi.org/#1}}\fi
\ifx \betal  \undefined \def \betal{\textit{et al.}}\fi
\ifx \binstitute  \undefined \def \binstitute#1{#1}\fi
\ifx \binstitutionaled  \undefined \def \binstitutionaled#1{#1}\fi
\ifx \bctitle  \undefined \def \bctitle#1{#1}\fi
\ifx \beditor  \undefined \def \beditor#1{#1}\fi
\ifx \bpublisher  \undefined \def \bpublisher#1{#1}\fi
\ifx \bbtitle  \undefined \def \bbtitle#1{#1}\fi
\ifx \bedition  \undefined \def \bedition#1{#1}\fi
\ifx \bseriesno  \undefined \def \bseriesno#1{#1}\fi
\ifx \blocation  \undefined \def \blocation#1{#1}\fi
\ifx \bsertitle  \undefined \def \bsertitle#1{#1}\fi
\ifx \bsnm \undefined \def \bsnm#1{#1}\fi
\ifx \bsuffix \undefined \def \bsuffix#1{#1}\fi
\ifx \bparticle \undefined \def \bparticle#1{#1}\fi
\ifx \barticle \undefined \def \barticle#1{#1}\fi
\bibcommenthead
\ifx \bconfdate \undefined \def \bconfdate #1{#1}\fi
\ifx \botherref \undefined \def \botherref #1{#1}\fi
\ifx \url \undefined \def \url#1{\textsf{#1}}\fi
\ifx \bchapter \undefined \def \bchapter#1{#1}\fi
\ifx \bbook \undefined \def \bbook#1{#1}\fi
\ifx \bcomment \undefined \def \bcomment#1{#1}\fi
\ifx \oauthor \undefined \def \oauthor#1{#1}\fi
\ifx \citeauthoryear \undefined \def \citeauthoryear#1{#1}\fi
\ifx \endbibitem  \undefined \def \endbibitem {}\fi
\ifx \bconflocation  \undefined \def \bconflocation#1{#1}\fi
\ifx \arxivurl  \undefined \def \arxivurl#1{\textsf{#1}}\fi
\csname PreBibitemsHook\endcsname

\bibitem{PPG2007Tech}
\begin{barticle}
\bauthor{\bsnm{Allen}, \binits{J.}}:
\batitle{Photoplethysmography and its application in clinical physiological
  measurement}.
\bjtitle{Physiological Measurement}
\bvolume{28},
\bfpage{1}--\blpage{39}
(\byear{2007}).
\doiurl{10.1088/0967-3334/28/3/R01}
\end{barticle}
\endbibitem

\bibitem{CurrentStatePPGs4Health}
\begin{botherref}
\oauthor{\bsnm{Tamura}, \binits{T.}}:
Current progress of photoplethysmography and spo2 for health monitoring.
Biomedical Engineering Letters
\textbf{9}
(2019).
\doiurl{10.1007/s13534-019-00097-w}
\end{botherref}
\endbibitem

\bibitem{PPG4SleepControl2017}
\begin{botherref}
\oauthor{\bsnm{Fonseca}, \binits{P.}},
\oauthor{\bsnm{Weysen}, \binits{T.}},
\oauthor{\bsnm{Goelema}, \binits{M.S.}},
\oauthor{\bsnm{Møst}, \binits{E.I.S.}},
\oauthor{\bsnm{Radha}, \binits{M.}},
\oauthor{\bsnm{Lunsingh~Scheurleer}, \binits{C.}},
\oauthor{\bparticle{van~den} \bsnm{Heuvel}, \binits{L.}},
\oauthor{\bsnm{Aarts}, \binits{R.M.}}:
{Validation of Photoplethysmography-Based Sleep Staging Compared With
  Polysomnography in Healthy Middle-Aged Adults}.
Sleep
\textbf{40}(7)
(2017).
\doiurl{10.1093/sleep/zsx097}
\end{botherref}
\endbibitem

\bibitem{OuraCVDs2020Kinnunen}
\begin{botherref}
\oauthor{\bsnm{Kinnunen}, \binits{H.}},
\oauthor{\bsnm{Rantanen}, \binits{A.}},
\oauthor{\bsnm{Kenttä}, \binits{T.}},
\oauthor{\bsnm{Koskimäki}, \binits{H.}}:
Feasible assessment of recovery and cardiovascular health: Accuracy of
  nocturnal hr and hrv assessed via ring ppg in comparison to medical grade
  ecg.
Physiological Measurement
\textbf{41}
(2020).
\doiurl{10.1088/1361-6579/ab840a}
\end{botherref}
\endbibitem

\bibitem{MeditationDetection2021Alvarez}
\begin{bchapter}
\bauthor{\bparticle{\'{A}lvarez} \bsnm{Casado}, \binits{C.}},
\bauthor{\bsnm{Paananen}, \binits{P.}},
\bauthor{\bsnm{Siirtola}, \binits{P.}},
\bauthor{\bsnm{Pirttikangas}, \binits{S.}},
\bauthor{\bsnm{Bordallo~L\'{o}pez}, \binits{M.}}:
\bctitle{Meditation detection using sensors from wearable devices}.
In: \bbtitle{Adjunct Proceedings of the International Joint Conference on
  Pervasive and Ubiquitous Computing},
pp. \bfpage{112}--\blpage{116}.
\bpublisher{ACM},
\blocation{New York, NY, USA}
(\byear{2021}).
\doiurl{10.1145/3460418.3479318}
\end{bchapter}
\endbibitem

\bibitem{PPGMonitoring2019SportPerformance}
\begin{botherref}
\oauthor{\bsnm{Seshadri}, \binits{D.R.}},
\oauthor{\bsnm{Li}, \binits{R.T.}},
\oauthor{\bsnm{Voos}, \binits{J.E.}},
\oauthor{\bsnm{Rowbottom}, \binits{J.R.}},
\oauthor{\bsnm{Alfes}, \binits{C.M.}},
\oauthor{\bsnm{Zorman}, \binits{C.A.}},
\oauthor{\bsnm{Drummond}, \binits{C.K.}}:
Wearable sensors for monitoring the internal and external workload of the
  athlete.
NPJ Digital Medicine
\textbf{2}
(2019)
\end{botherref}
\endbibitem

\bibitem{rPPGFundaments2015}
\begin{botherref}
\oauthor{\bsnm{Sun}, \binits{Y.}},
\oauthor{\bsnm{Thakor}, \binits{N.}}:
Photoplethysmography revisited: From contact to noncontact, from point to
  imaging.
IEEE transactions on bio-medical engineering
\textbf{63}
(2015).
\doiurl{10.1109/TBME.2015.2476337}
\end{botherref}
\endbibitem

\bibitem{RecentReviewRPPGMethods2021}
\begin{barticle}
\bauthor{\bsnm{Ni}, \binits{A.}},
\bauthor{\bsnm{Azarang}, \binits{A.}},
\bauthor{\bsnm{Kehtarnavaz}, \binits{N.}}:
\batitle{A review of deep learning-based contactless heart rate measurement
  methods}.
\bjtitle{Sensors}
\bvolume{21},
\bfpage{3719}
(\byear{2021}).
\doiurl{10.3390/s21113719}
\end{barticle}
\endbibitem

\bibitem{RemoteMonitoringHealth2019Review}
\begin{botherref}
\oauthor{\bsnm{Khanam}, \binits{F.-T.-Z.}},
\oauthor{\bsnm{Al-Naji}, \binits{A.}},
\oauthor{\bsnm{Chahl}, \binits{J.}}:
Remote monitoring of vital signs in diverse non-clinical and clinical scenarios
  using computer vision systems: A review.
Applied Sciences
\textbf{9}(20)
(2019).
\doiurl{10.3390/app9204474}
\end{botherref}
\endbibitem

\bibitem{GreenMethod2008}
\begin{barticle}
\bauthor{\bsnm{Verkruysse}, \binits{W.}},
\bauthor{\bsnm{Svaasand}, \binits{L.O.}},
\bauthor{\bsnm{Nelson}, \binits{J.S.}}:
\batitle{Remote plethysmographic imaging using ambient light.}
\bjtitle{Opt. Express}
\bvolume{16}(\bissue{26}),
\bfpage{21434}--\blpage{21445}
(\byear{2008}).
\doiurl{10.1364/OE.16.021434}
\end{barticle}
\endbibitem

\bibitem{PohMethod2011}
\begin{barticle}
\bauthor{\bsnm{Poh}, \binits{M.-Z.}},
\bauthor{\bsnm{McDuff}, \binits{D.J.}},
\bauthor{\bsnm{Picard}, \binits{R.W.}}:
\batitle{Advancements in noncontact, multiparameter physiological measurements
  using a webcam}.
\bjtitle{IEEE Tran. on Biomedical Engineering}
\bvolume{58}(\bissue{1}),
\bfpage{7}--\blpage{11}
(\byear{2011}).
\doiurl{10.1109/TBME.2010.2086456}
\end{barticle}
\endbibitem

\bibitem{PCAMethod2011}
\begin{bchapter}
\bauthor{\bsnm{Lewandowska}, \binits{M.}},
\bauthor{\bsnm{Rumiński}, \binits{J.}},
\bauthor{\bsnm{Kocejko}, \binits{T.}},
\bauthor{\bsnm{Nowak}, \binits{J.}}:
\bctitle{Measuring pulse rate with a webcam — a non-contact method for
  evaluating cardiac activity}.
In: \bbtitle{Federated Conference on Computer Science and Information Systems
  (FedCSIS)},
pp. \bfpage{405}--\blpage{410}
(\byear{2011})
\end{bchapter}
\endbibitem

\bibitem{CHROMMethod2013}
\begin{barticle}
\bauthor{\bparticle{de} \bsnm{Haan}, \binits{G.}},
\bauthor{\bsnm{Jeanne}, \binits{V.}}:
\batitle{Robust pulse rate from chrominance-based rppg}.
\bjtitle{IEEE Transactions on Biomedical Engineering}
\bvolume{60}(\bissue{10}),
\bfpage{2878}--\blpage{2886}
(\byear{2013}).
\doiurl{10.1109/TBME.2013.2266196}
\end{barticle}
\endbibitem

\bibitem{PVBMethod2014}
\begin{barticle}
\bauthor{\bsnm{Haan}, \binits{G.}},
\bauthor{\bsnm{Leest}, \binits{A.}}:
\batitle{Improved motion robustness of remote-ppg by using the blood volume
  pulse signature}.
\bjtitle{Physiological measurement}
\bvolume{35},
\bfpage{1913}--\blpage{1926}
(\byear{2014}).
\doiurl{10.1088/0967-3334/35/9/1913}
\end{barticle}
\endbibitem

\bibitem{LiMethod2014}
\begin{bchapter}
\bauthor{\bsnm{Li}, \binits{X.}},
\bauthor{\bsnm{Chen}, \binits{J.}},
\bauthor{\bsnm{Zhao}, \binits{G.}},
\bauthor{\bsnm{Pietikäinen}, \binits{M.}}:
\bctitle{Remote heart rate measurement from face videos under realistic
  situations}.
In: \bbtitle{IEEE Conf on Computer Vision and Pattern Recognition},
pp. \bfpage{4264}--\blpage{4271}
(\byear{2014}).
\doiurl{10.1109/CVPR.2014.543}
\end{bchapter}
\endbibitem

\bibitem{2SRMethod2016}
\begin{barticle}
\bauthor{\bsnm{Wang}, \binits{W.}},
\bauthor{\bsnm{Stuijk}, \binits{S.}},
\bauthor{\bparticle{de} \bsnm{Haan}, \binits{G.}}:
\batitle{A novel algorithm for remote photoplethysmography: Spatial subspace
  rotation}.
\bjtitle{IEEE Transactions on Biomedical Engineering}
\bvolume{63}(\bissue{9}),
\bfpage{1974}--\blpage{1984}
(\byear{2016}).
\doiurl{10.1109/TBME.2015.2508602}
\end{barticle}
\endbibitem

\bibitem{Lab2016SNR}
\begin{barticle}
\bauthor{\bsnm{Yang}, \binits{Y.}},
\bauthor{\bsnm{Liu}, \binits{C.}},
\bauthor{\bsnm{Yu}, \binits{H.}},
\bauthor{\bsnm{Shao}, \binits{D.}},
\bauthor{\bsnm{Tsow}, \binits{F.}},
\bauthor{\bsnm{Tao}, \binits{N.}}:
\batitle{Motion robust remote photoplethysmography in cielab color space}.
\bjtitle{Journal of Biomedical Optics}
\bvolume{21},
\bfpage{117001}
(\byear{2016}).
\doiurl{10.1117/1.JBO.21.11.117001}
\end{barticle}
\endbibitem

\bibitem{POSMethod2017}
\begin{barticle}
\bauthor{\bsnm{Wang}, \binits{W.}},
\bauthor{\bparticle{den} \bsnm{Brinker}, \binits{A.C.}},
\bauthor{\bsnm{Stuijk}, \binits{S.}},
\bauthor{\bparticle{de} \bsnm{Haan}, \binits{G.}}:
\batitle{Algorithmic principles of remote ppg}.
\bjtitle{IEEE Transactions on Biomedical Engineering}
\bvolume{64}(\bissue{7}),
\bfpage{1479}--\blpage{1491}
(\byear{2017}).
\doiurl{10.1109/TBME.2016.2609282}
\end{barticle}
\endbibitem

\bibitem{LGIMethod2018}
\begin{bchapter}
\bauthor{\bsnm{Pilz}, \binits{C.S.}},
\bauthor{\bsnm{Zaunseder}, \binits{S.}},
\bauthor{\bsnm{Krajewski}, \binits{J.}},
\bauthor{\bsnm{Blazek}, \binits{V.}}:
\bctitle{Local group invariance for heart rate estimation from face videos in
  the wild}.
In: \bbtitle{2018 IEEE/CVF Conference on Computer Vision and Pattern
  Recognition Workshops (CVPRW)},
pp. \bfpage{1335}--\blpage{13358}
(\byear{2018}).
\doiurl{10.1109/CVPRW.2018.00172}
\end{bchapter}
\endbibitem

\bibitem{FaceRPPG2020Gudi}
\begin{botherref}
\oauthor{\bsnm{Gudi}, \binits{A.}},
\oauthor{\bsnm{Bittner}, \binits{M.}},
\oauthor{\bparticle{van} \bsnm{Gemert}, \binits{J.}}:
Real-time webcam heart-rate and variability estimation with clean ground truth
  for evaluation.
Applied Sciences
\textbf{10}(23)
(2020).
\doiurl{10.3390/app10238630}
\end{botherref}
\endbibitem

\bibitem{gideon2021way}
\begin{bchapter}
\bauthor{\bsnm{Gideon}, \binits{J.}},
\bauthor{\bsnm{Stent}, \binits{S.}}:
\bctitle{The way to my heart is through contrastive learning: Remote
  photoplethysmography from unlabelled video}.
In: \bbtitle{Proceedings of the IEEE/CVF International Conference on Computer
  Vision},
pp. \bfpage{3995}--\blpage{4004}
(\byear{2021})
\end{bchapter}
\endbibitem

\bibitem{sun2022contrast}
\begin{bchapter}
\bauthor{\bsnm{Sun}, \binits{Z.}},
\bauthor{\bsnm{Li}, \binits{X.}}:
\bctitle{Contrast-phys: Unsupervised video-based remote physiological
  measurement via spatiotemporal contrast}.
In: \bbtitle{Computer Vision--ECCV 2022: 17th European Conference, Tel Aviv,
  Israel, October 23--27, 2022, Proceedings, Part XII},
pp. \bfpage{492}--\blpage{510}
(\byear{2022}).
\bcomment{Springer}
\end{bchapter}
\endbibitem

\bibitem{Hsu2014SVMHR}
\begin{bchapter}
\bauthor{\bsnm{Hsu}, \binits{Y.}},
\bauthor{\bsnm{Lin}, \binits{Y.-L.}},
\bauthor{\bsnm{Hsu}, \binits{W.}}:
\bctitle{Learning-based heart rate detection from remote photoplethysmography
  features},
pp. \bfpage{4433}--\blpage{4437}
(\byear{2014}).
\doiurl{10.1109/ICASSP.2014.6854440}
\end{bchapter}
\endbibitem

\bibitem{Fan2015BayesHR}
\begin{bchapter}
\bauthor{\bsnm{Fan}, \binits{X.}},
\bauthor{\bsnm{Wang}, \binits{J.}}:
\bctitle{Bayesheart: A probabilistic approach for robust, low-latency heart
  rate monitoring on camera phones}.
In: \bbtitle{Proceedings of the 20th International Conference on Intelligent
  User Interfaces}.
\bsertitle{IUI '15},
pp. \bfpage{405}--\blpage{416}.
\bpublisher{Association for Computing Machinery},
\blocation{New York, NY, USA}
(\byear{2015}).
\doiurl{10.1145/2678025.2701364}
\end{bchapter}
\endbibitem

\bibitem{HRCNNMethod2018}
\begin{bchapter}
\bauthor{\bsnm{{\v{S}}petl{\'\i}k}, \binits{R.}},
\bauthor{\bsnm{Franc}, \binits{V.}},
\bauthor{\bsnm{Matas}, \binits{J.}}:
\bctitle{Visual heart rate estimation with convolutional neural network}.
In: \bbtitle{Proceedings of the British Machine Vision Conference, Newcastle,
  UK},
pp. \bfpage{3}--\blpage{6}
(\byear{2018})
\end{bchapter}
\endbibitem

\bibitem{DeepPhysMethod2018}
\begin{botherref}
\oauthor{\bsnm{Chen}, \binits{W.V.}},
\oauthor{\bsnm{McDuff}, \binits{D.J.}}:
Deepphys: Video-based physiological measurement using convolutional attention
  networks.
ArXiv
\textbf{abs/1805.07888}
(2018)
\end{botherref}
\endbibitem

\bibitem{RhythmNet2019}
\begin{barticle}
\bauthor{\bsnm{Niu}, \binits{X.}},
\bauthor{\bsnm{Shan}, \binits{S.}},
\bauthor{\bsnm{Han}, \binits{H.}},
\bauthor{\bsnm{Chen}, \binits{X.}}:
\batitle{Rhythmnet: End-to-end heart rate estimation from face via
  spatial-temporal representation}.
\bjtitle{IEEE Transactions on Image Processing}
\bvolume{29},
\bfpage{2409}--\blpage{2423}
(\byear{2019})
\end{barticle}
\endbibitem

\bibitem{ZitongMethod2019}
\begin{bchapter}
\bauthor{\bsnm{Yu}, \binits{Z.}},
\bauthor{\bsnm{Peng}, \binits{W.}},
\bauthor{\bsnm{Li}, \binits{X.}},
\bauthor{\bsnm{Hong}, \binits{X.}},
\bauthor{\bsnm{Zhao}, \binits{G.}}:
\bctitle{Remote heart rate measurement from highly compressed facial videos: An
  end-to-end deep learning solution with video enhancement}.
In: \bbtitle{IEEE/CVF International Conference on Computer Vision (ICCV)},
pp. \bfpage{151}--\blpage{160}
(\byear{2019}).
\doiurl{10.1109/ICCV.2019.00024}
\end{bchapter}
\endbibitem

\bibitem{AutoHR2020Zitong}
\begin{barticle}
\bauthor{\bsnm{Yu}, \binits{Z.}},
\bauthor{\bsnm{Li}, \binits{X.}},
\bauthor{\bsnm{Niu}, \binits{X.}},
\bauthor{\bsnm{Shi}, \binits{J.}},
\bauthor{\bsnm{Zhao}, \binits{G.}}:
\batitle{Autohr: A strong end-to-end baseline for remote heart rate measurement
  with neural searching}.
\bjtitle{IEEE Signal Processing Letters}
\bvolume{27},
\bfpage{1245}--\blpage{1249}
(\byear{2020})
\end{barticle}
\endbibitem

\bibitem{MetaRPPG2020}
\begin{bchapter}
\bauthor{\bsnm{Lee}, \binits{E.}},
\bauthor{\bsnm{Chen}, \binits{E.}},
\bauthor{\bsnm{Lee}, \binits{C.-Y.}}:
\bctitle{Meta-rppg: Remote heart rate estimation using a transductive
  meta-learner}.
In: \beditor{\bsnm{Vedaldi}, \binits{A.}},
\beditor{\bsnm{Bischof}, \binits{H.}},
\beditor{\bsnm{Brox}, \binits{T.}},
\beditor{\bsnm{Frahm}, \binits{J.-M.}} (eds.)
\bbtitle{Computer Vision -- ECCV 2020},
pp. \bfpage{392}--\blpage{409}.
\bpublisher{Springer},
\blocation{Cham}
(\byear{2020})
\end{bchapter}
\endbibitem

\bibitem{PulseGAN2021}
\begin{barticle}
\bauthor{\bsnm{Song}, \binits{R.}},
\bauthor{\bsnm{Chen}, \binits{H.}},
\bauthor{\bsnm{Cheng}, \binits{J.}},
\bauthor{\bsnm{Li}, \binits{C.}},
\bauthor{\bsnm{Liu}, \binits{Y.}},
\bauthor{\bsnm{Chen}, \binits{X.}}:
\batitle{Pulsegan: Learning to generate realistic pulse waveforms in remote
  photoplethysmography}.
\bjtitle{IEEE Journal of Biomedical and Health Informatics}
\bvolume{PP},
\bfpage{1}--\blpage{1}
(\byear{2021}).
\doiurl{10.1109/JBHI.2021.3051176}
\end{barticle}
\endbibitem

\bibitem{ANDrPPG2022}
\begin{barticle}
\bauthor{\bsnm{Lokendra}, \binits{B.}},
\bauthor{\bsnm{Puneet}, \binits{G.}}:
\batitle{And-rppg: A novel denoising-rppg network for improving remote heart
  rate estimation}.
\bjtitle{Computers in Biology and Medicine}
\bvolume{141},
\bfpage{105146}
(\byear{2022})
\end{barticle}
\endbibitem

\bibitem{Survey_DL_RPPG2021_Cheng}
\begin{botherref}
\oauthor{\bsnm{Cheng}, \binits{C.-H.}},
\oauthor{\bsnm{Wong}, \binits{K.-L.}},
\oauthor{\bsnm{Chin}, \binits{J.-W.}},
\oauthor{\bsnm{Chan}, \binits{T.-T.}},
\oauthor{\bsnm{So}, \binits{R.H.Y.}}:
Deep learning methods for remote heart rate measurement: A review and future
  research agenda.
Sensors
\textbf{21}(18)
(2021).
\doiurl{10.3390/s21186296}
\end{botherref}
\endbibitem

\bibitem{CNNrPPG_Limitations2020}
\begin{barticle}
\bauthor{\bsnm{Zhan}, \binits{Q.}},
\bauthor{\bsnm{Wang}, \binits{W.}},
\bauthor{\bparticle{de} \bsnm{Haan}, \binits{G.}}:
\batitle{Analysis of cnn-based remote-ppg to understand limitations and
  sensitivities}.
\bjtitle{Biomed. Opt. Express}
\bvolume{11}(\bissue{3}),
\bfpage{1268}--\blpage{1283}
(\byear{2020}).
\doiurl{10.1364/BOE.382637}
\end{barticle}
\endbibitem

\bibitem{pyVHR2020}
\begin{barticle}
\bauthor{\bsnm{Boccignone}, \binits{G.}},
\bauthor{\bsnm{Conte}, \binits{D.}},
\bauthor{\bsnm{Cuculo}, \binits{V.}},
\bauthor{\bsnm{D’Amelio}, \binits{A.}},
\bauthor{\bsnm{Grossi}, \binits{G.}},
\bauthor{\bsnm{Lanzarotti}, \binits{R.}}:
\batitle{An open framework for remote-ppg methods and their assessment}.
\bjtitle{IEEE Access}
\bvolume{8},
\bfpage{216083}--\blpage{216103}
(\byear{2020}).
\doiurl{10.1109/ACCESS.2020.3040936}
\end{barticle}
\endbibitem

\bibitem{MTCNN2016}
\begin{barticle}
\bauthor{\bsnm{Zhang}, \binits{K.}},
\bauthor{\bsnm{Zhang}, \binits{Z.}},
\bauthor{\bsnm{Li}, \binits{Z.}},
\bauthor{\bsnm{Qiao}, \binits{Y.}}:
\batitle{Joint face detection and alignment using multitask cascaded
  convolutional networks}.
\bjtitle{IEEE Signal Processing Letters}
\bvolume{23}(\bissue{10}),
\bfpage{1499}--\blpage{1503}
(\byear{2016}).
\doiurl{10.1109/lsp.2016.2603342}
\end{barticle}
\endbibitem

\bibitem{dlib09}
\begin{barticle}
\bauthor{\bsnm{King}, \binits{D.E.}}:
\batitle{Dlib-ml: A machine learning toolkit}.
\bjtitle{Journal of Machine Learning Research}
\bvolume{10},
\bfpage{1755}--\blpage{1758}
(\byear{2009})
\end{barticle}
\endbibitem

\bibitem{SSDFaceDetector2015}
\begin{bchapter}
\bauthor{\bsnm{Liu}, \binits{W.}},
\bauthor{\bsnm{Anguelov}, \binits{D.}},
\bauthor{\bsnm{Erhan}, \binits{D.}},
\bauthor{\bsnm{Szegedy}, \binits{C.}},
\bauthor{\bsnm{Reed}, \binits{S.}},
\bauthor{\bsnm{Fu}, \binits{C.-Y.}},
\bauthor{\bsnm{Berg}, \binits{A.C.}}:
\bctitle{Ssd: Single shot multibox detector}.
In: \bbtitle{European Conference on Computer Vision},
pp. \bfpage{21}--\blpage{37}
(\byear{2016}).
\bcomment{Springer}
\end{bchapter}
\endbibitem

\bibitem{SSDvsMTCNN2021}
\begin{botherref}
\oauthor{\bsnm{Sanchez-Moreno}, \binits{A.S.}},
\oauthor{\bsnm{Olivares-Mercado}, \binits{J.}},
\oauthor{\bsnm{Hernandez-Suarez}, \binits{A.}},
\oauthor{\bsnm{Toscano-Medina}, \binits{K.}},
\oauthor{\bsnm{Sanchez-Perez}, \binits{G.}},
\oauthor{\bsnm{Benitez-Garcia}, \binits{G.}}:
Efficient face recognition system for operating in unconstrained environments.
Journal of Imaging
\textbf{7}(9)
(2021).
\doiurl{10.3390/jimaging7090161}
\end{botherref}
\endbibitem

\bibitem{ERTFace2014}
\begin{bchapter}
\bauthor{\bsnm{{Kazemi}}, \binits{V.}},
\bauthor{\bsnm{{Sullivan}}, \binits{J.}}:
\bctitle{One millisecond face alignment with an ensemble of regression trees}.
In: \bbtitle{2014 IEEE Conference on Computer Vision and Pattern Recognition},
pp. \bfpage{1867}--\blpage{1874}
(\byear{2014}).
\doiurl{10.1109/CVPR.2014.241}
\end{bchapter}
\endbibitem

\bibitem{DANKowalskiNT17}
\begin{bchapter}
\bauthor{\bsnm{Kowalski}, \binits{M.}},
\bauthor{\bsnm{Naruniec}, \binits{J.}},
\bauthor{\bsnm{Trzcinski}, \binits{T.}}:
\bctitle{Deep alignment network: A convolutional neural network for robust face
  alignment}.
In: \bbtitle{Proceedings of the IEEE Conference on Computer Vision and Pattern
  Recognition Workshops},
pp. \bfpage{88}--\blpage{97}
(\byear{2017})
\end{bchapter}
\endbibitem

\bibitem{AlvarezBordallo2021FaceAlignment}
\begin{botherref}
\oauthor{\bsnm{{\'A}lvarez~Casado}, \binits{C.}},
\oauthor{\bsnm{Bordallo~L{\'o}pez}, \binits{M.}}:
Real-time face alignment: evaluation methods, training strategies and
  implementation optimization.
Journal of Real-Time Image Processing,
1--29
(2021)
\end{botherref}
\endbibitem

\bibitem{KaiserWindow2019}
\begin{bchapter}
\bauthor{\bsnm{Rakshit}, \binits{H.}},
\bauthor{\bsnm{Ullah}, \binits{M.A.}}:
\bctitle{A comparative study on window functions for designing efficient fir
  filter}.
In: \bbtitle{2014 9th International Forum on Strategic Technology (IFOST)},
pp. \bfpage{91}--\blpage{96}
(\byear{2014}).
\doiurl{10.1109/IFOST.2014.6991079}
\end{bchapter}
\endbibitem

\bibitem{Unakafov_2018}
\begin{barticle}
\bauthor{\bsnm{Unakafov}, \binits{A.M.}}:
\batitle{Pulse rate estimation using imaging photoplethysmography: generic
  framework and comparison of methods on a publicly available dataset}.
\bjtitle{Biomedical Physics {\&} Engineering Express}
\bvolume{4}(\bissue{4}),
\bfpage{045001}
(\byear{2018}).
\doiurl{10.1088/2057-1976/aabd09}
\end{barticle}
\endbibitem

\bibitem{LABSkinColor2020}
\begin{barticle}
\bauthor{\bsnm{Ly}, \binits{B.}},
\bauthor{\bsnm{Dyer}, \binits{E.}},
\bauthor{\bsnm{Feig}, \binits{J.}},
\bauthor{\bsnm{Chien}, \binits{A.}},
\bauthor{\bsnm{Bino}, \binits{S.}}:
\batitle{Research techniques made simple: Cutaneous colorimetry: A reliable
  technique for objective skin color measurement}.
\bjtitle{The Journal of investigative dermatology}
\bvolume{140},
\bfpage{3}--\blpage{121}
(\byear{2020}).
\doiurl{10.1016/j.jid.2019.11.003}
\end{barticle}
\endbibitem

\bibitem{PPGFingerPressureChanges}
\begin{barticle}
\bauthor{\bsnm{Teng}, \binits{X.-F.}},
\bauthor{\bsnm{Zhang}, \binits{Y.-T.}}:
\batitle{Theoretical study on the effect of sensor contact force on pulse
  transit time}.
\bjtitle{IEEE Transactions on Biomedical Engineering}
\bvolume{54}(\bissue{8}),
\bfpage{1490}--\blpage{1498}
(\byear{2007}).
\doiurl{10.1109/TBME.2007.900815}
\end{barticle}
\endbibitem

\bibitem{FilteringEffectTime2021}
\begin{barticle}
\bauthor{\bsnm{Liu}, \binits{H.}},
\bauthor{\bsnm{Allen}, \binits{J.}},
\bauthor{\bsnm{Khalid}, \binits{S.G.}},
\bauthor{\bsnm{Chen}, \binits{F.}},
\bauthor{\bsnm{Zheng}, \binits{D.}}:
\batitle{Filtering-induced time shifts in photoplethysmography pulse features
  measured at different body sites: the importance of filter definition and
  standardization}.
\bjtitle{Physiological Measurement}
\bvolume{42}(\bissue{7}),
\bfpage{074001}
(\byear{2021}).
\doiurl{10.1088/1361-6579/ac0a34}
\end{barticle}
\endbibitem

\bibitem{InaccuracySources2021PPG}
\begin{botherref}
\oauthor{\bsnm{Fine}, \binits{J.}},
\oauthor{\bsnm{Branan}, \binits{K.L.}},
\oauthor{\bsnm{Rodriguez}, \binits{A.J.}},
\oauthor{\bsnm{Boonya-ananta}, \binits{T.}},
\oauthor{\bsnm{Ajmal}},
\oauthor{\bsnm{Ramella-Roman}, \binits{J.C.}},
\oauthor{\bsnm{McShane}, \binits{M.J.}},
\oauthor{\bsnm{Coté}, \binits{G.L.}}:
Sources of inaccuracy in photoplethysmography for continuous cardiovascular
  monitoring.
Biosensors
\textbf{11}(4)
(2021).
\doiurl{10.3390/bios11040126}
\end{botherref}
\endbibitem

\bibitem{respiratoryPPG2019EffectsSite}
\begin{barticle}
\bauthor{\bsnm{Hartmann}, \binits{V.}},
\bauthor{\bsnm{Liu}, \binits{H.}},
\bauthor{\bsnm{Chen}, \binits{F.}},
\bauthor{\bsnm{Hong}, \binits{W.}},
\bauthor{\bsnm{Hughes}, \binits{S.}},
\bauthor{\bsnm{Zheng}, \binits{D.}}:
\batitle{Toward accurate extraction of respiratory frequency from the
  photoplethysmogram: Effect of measurement site}.
\bjtitle{Frontiers in Physiology}
\bvolume{10},
\bfpage{732}
(\byear{2019}).
\doiurl{10.3389/fphys.2019.00732}
\end{barticle}
\endbibitem

\bibitem{BloodPerfusionDifferences2018}
\begin{barticle}
\bauthor{\bsnm{Rodríguez}, \binits{A.}},
\bauthor{\bsnm{Ramos}, \binits{J.}}:
\batitle{Video pulse rate variability analysis in stationary and motion
  conditions}.
\bjtitle{Biomedical engineering online}
\bvolume{17},
\bfpage{11}
(\byear{2018}).
\doiurl{10.1186/s12938-018-0437-0}
\end{barticle}
\endbibitem

\bibitem{Nanni2023SkinDetection}
\begin{barticle}
\bauthor{\bsnm{Nanni}, \binits{L.}},
\bauthor{\bsnm{Loreggia}, \binits{A.}},
\bauthor{\bsnm{Lumini}, \binits{A.}},
\bauthor{\bsnm{Dorizza}, \binits{A.}}:
\batitle{A standardized approach for skin detection: Analysis of the literature
  and case studies}.
\bjtitle{Journal of Imaging}
\bvolume{9},
\bfpage{35}
(\byear{2023}).
\doiurl{10.3390/jimaging9020035}
\end{barticle}
\endbibitem

\bibitem{FractalSeries2002}
\begin{barticle}
\bauthor{\bsnm{Eke}, \binits{A.}},
\bauthor{\bsnm{Herman}, \binits{P.}},
\bauthor{\bsnm{Kocsis}, \binits{L.}},
\bauthor{\bsnm{Kozák}, \binits{L.}}:
\batitle{Fractal characterization of complexity in physiological temporal
  signals}.
\bjtitle{Physiol Meas}
\bvolume{23},
\bfpage{31}--\blpage{38}
(\byear{2002})
\end{barticle}
\endbibitem

\bibitem{DFAPhysiological1995}
\begin{barticle}
\bauthor{\bsnm{Peng}, \binits{C.-K.}},
\bauthor{\bsnm{Havlin}, \binits{S.}},
\bauthor{\bsnm{Stanley}, \binits{H.}},
\bauthor{\bsnm{Goldberger}, \binits{A.}}:
\batitle{Quantification of scaling exponents and crossover phenomena in
  nonstationary heartbeat time series}.
\bjtitle{Chaos: An Interdisciplinary Journal of Nonlinear Science}
\bvolume{5},
\bfpage{82}
(\byear{1995})
\end{barticle}
\endbibitem

\bibitem{Physionet_MIT_ToolKit2000}
\begin{barticle}
\bauthor{\bsnm{Goldberger}, \binits{A.L.}},
\bauthor{\bsnm{Amaral}, \binits{L.A.N.}},
\bauthor{\bsnm{Glass}, \binits{L.}},
\bauthor{\bsnm{Hausdorff}, \binits{J.M.}},
\bauthor{\bsnm{Ivanov}, \binits{P.C.}},
\bauthor{\bsnm{Mark}, \binits{R.G.}},
\bauthor{\bsnm{Mietus}, \binits{J.E.}},
\bauthor{\bsnm{Moody}, \binits{G.B.}},
\bauthor{\bsnm{Peng}, \binits{C.-K.}},
\bauthor{\bsnm{Stanley}, \binits{H.E.}}:
\batitle{Physiobank, physiotoolkit, and physionet}.
\bjtitle{Circulation}
\bvolume{101}(\bissue{23}),
\bfpage{215}--\blpage{220}
(\byear{2000}).
\doiurl{10.1161/01.CIR.101.23.e215}
\end{barticle}
\endbibitem

\bibitem{stoica2005spectral}
\begin{botherref}
\oauthor{\bsnm{Stoica}, \binits{P.}}, et al.:
Spectral Analysis of Signals
vol. 452
\end{botherref}
\endbibitem

\bibitem{Biosignals2005SpectralAnalysis}
\begin{bbook}
\bauthor{\bsnm{Sörnmo}, \binits{L.}},
\bauthor{\bsnm{Laguna}, \binits{P.}}:
\bbtitle{Bioelectrical Signal Processing in Cardiac and Neurological
  Applications},
(\byear{2005}).
\doiurl{10.1016/B978-0-12-437552-9.X5000-4}
\end{bbook}
\endbibitem

\bibitem{golub2013matrix}
\begin{bbook}
\bauthor{\bsnm{Golub}, \binits{G.H.}},
\bauthor{\bsnm{Van~Loan}, \binits{C.F.}}:
\bbtitle{Matrix Computations}.
\bpublisher{JHU press}, \blocation{???}
(\byear{2013})
\end{bbook}
\endbibitem

\bibitem{HouseholderReflections1958}
\begin{barticle}
\bauthor{\bsnm{Householder}, \binits{A.S.}}:
\batitle{Unitary triangularization of a nonsymmetric matrix}.
\bjtitle{Journal of the ACM (JACM)}
\bvolume{5}(\bissue{4}),
\bfpage{339}--\blpage{342}
(\byear{1958})
\end{barticle}
\endbibitem

\bibitem{shao2021householder}
\begin{botherref}
\oauthor{\bsnm{Shao}, \binits{M.}}:
Householder orthogonalization with a non-standard inner product.
arXiv preprint arXiv:2104.04180
(2021)
\end{botherref}
\endbibitem

\bibitem{QR_RecoverCorruptedImages2019}
\begin{barticle}
\bauthor{\bsnm{Liu}, \binits{Q.}},
\bauthor{\bsnm{Davoine}, \binits{F.}},
\bauthor{\bsnm{Yang}, \binits{J.}},
\bauthor{\bsnm{Cui}, \binits{Y.}},
\bauthor{\bsnm{Jin}, \binits{Z.}},
\bauthor{\bsnm{Han}, \binits{F.}}:
\batitle{A fast and accurate matrix completion method based on qr decomposition
  and l2,1-norm minimization}.
\bjtitle{IEEE Transactions on Neural Networks and Learning Systems}
\bvolume{30}(\bissue{3}),
\bfpage{803}--\blpage{817}
(\byear{2019}).
\doiurl{10.1109/TNNLS.2018.2851957}
\end{barticle}
\endbibitem

\bibitem{DenoisingQRD2016}
\begin{barticle}
\bauthor{\bsnm{Zimoń}, \binits{M.J.}},
\bauthor{\bsnm{Prosser}, \binits{R.}},
\bauthor{\bsnm{Emerson}, \binits{D.R.}},
\bauthor{\bsnm{Borg}, \binits{M.K.}},
\bauthor{\bsnm{Bray}, \binits{D.J.}},
\bauthor{\bsnm{Grinberg}, \binits{L.}},
\bauthor{\bsnm{Reese}, \binits{J.M.}}:
\batitle{An evaluation of noise reduction algorithms for particle-based fluid
  simulations in multi-scale applications}.
\bjtitle{Journal of Computational Physics}
\bvolume{325},
\bfpage{380}--\blpage{394}
(\byear{2016})
\end{barticle}
\endbibitem

\bibitem{QRFactorization1992Application}
\begin{barticle}
\bauthor{\bsnm{Anderson}, \binits{E.}},
\bauthor{\bsnm{Bai}, \binits{Z.}},
\bauthor{\bsnm{Dongarra}, \binits{J.}}:
\batitle{Generalized qr factorization and its applications}.
\bjtitle{Linear Algebra and its Applications}
\bvolume{162-164},
\bfpage{243}--\blpage{271}
(\byear{1992})
\end{barticle}
\endbibitem

\bibitem{Sharma2013QR}
\begin{botherref}
\oauthor{\bsnm{Sharma}, \binits{A.}},
\oauthor{\bsnm{Paliwal}, \binits{K.}},
\oauthor{\bsnm{Imoto}, \binits{S.}},
\oauthor{\bsnm{Miyano}, \binits{S.}}:
Principal component analysis using qr decomposition.
International Journal of Machine Learning and Cybernetics
\textbf{4}
(2013).
\doiurl{10.1007/s13042-012-0131-7}
\end{botherref}
\endbibitem

\bibitem{CHEN2012QRSIMOvsSVD}
\begin{barticle}
\bauthor{\bsnm{Chen}, \binits{C.Y.}},
\bauthor{\bsnm{Yu}, \binits{J.S.}}:
\batitle{Using qr factorization in subspace based blind channel
  identification}.
\bjtitle{Procedia Engineering}
\bvolume{29},
\bfpage{3413}--\blpage{3417}
(\byear{2012}).
\doiurl{10.1016/j.proeng.2012.01.504}.
\bcomment{2012 International Workshop on Information and Electronics
  Engineering}
\end{barticle}
\endbibitem

\bibitem{Park2021QRMIMO}
\begin{botherref}
\oauthor{\bsnm{Park}, \binits{S.}}:
Scheduled qr-bp detector with interference cancellation and candidate
  constraints for mimo systems.
Sensors
\textbf{21}(11)
(2021).
\doiurl{10.3390/s21113734}
\end{botherref}
\endbibitem

\bibitem{Barnov2017QRBeamformingIntel}
\begin{bchapter}
\bauthor{\bsnm{Barnov}, \binits{A.}},
\bauthor{\bsnm{Bracha}, \binits{V.}},
\bauthor{\bsnm{Markovich-Golan}, \binits{S.}}:
\bctitle{Qrd based mvdr beamforming for fast tracking of speech and noise
  dynamics},
pp. \bfpage{369}--\blpage{373}
(\byear{2017}).
\doiurl{10.1109/WASPAA.2017.8170057}
\end{bchapter}
\endbibitem

\bibitem{Amintoosi2007QRBackgroundSub}
\begin{bchapter}
\bauthor{\bsnm{Amintoosi}, \binits{M.}},
\bauthor{\bsnm{Farbiz}, \binits{F.}},
\bauthor{\bsnm{Fathy}, \binits{M.}},
\bauthor{\bsnm{Analoui}, \binits{M.}},
\bauthor{\bsnm{Mozayani}, \binits{N.}}:
\bctitle{Qr decomposition-based algorithm for background subtraction},
vol. \bseriesno{1},
pp. \bfpage{1093}--\blpage{1096}
(\byear{2007}).
\doiurl{10.1109/ICASSP.2007.366102}
\end{bchapter}
\endbibitem

\bibitem{roberts2020qrBackPropagation}
\begin{botherref}
\oauthor{\bsnm{Roberts}, \binits{D.A.}},
\oauthor{\bsnm{Roberts}, \binits{L.R.}}:
Qr and lq decomposition matrix backpropagation algorithms for square, wide, and
  deep--real or complex--matrices and their software implementation.
arXiv preprint arXiv:2009.10071
(2020)
\end{botherref}
\endbibitem

\bibitem{LAPACKLib}
\begin{bchapter}
\bauthor{\bsnm{Angerson}, \binits{E.}},
\bauthor{\bsnm{Bai}, \binits{Z.}},
\bauthor{\bsnm{Dongarra}, \binits{J.}},
\bauthor{\bsnm{Greenbaum}, \binits{A.}},
\bauthor{\bsnm{McKenney}, \binits{A.}},
\bauthor{\bsnm{Du~Croz}, \binits{J.}},
\bauthor{\bsnm{Hammarling}, \binits{S.}},
\bauthor{\bsnm{Demmel}, \binits{J.}},
\bauthor{\bsnm{Bischof}, \binits{C.}},
\bauthor{\bsnm{Sorensen}, \binits{D.}}:
\bctitle{Lapack: A portable linear algebra library for high-performance
  computers}.
In: \bbtitle{Supercomputing '90:Proceedings of the 1990 ACM/IEEE Conference on
  Supercomputing},
pp. \bfpage{2}--\blpage{11}
(\byear{1990}).
\doiurl{10.1109/SUPERC.1990.129995}
\end{bchapter}
\endbibitem

\bibitem{MKLIntel}
\begin{botherref}
\oauthor{\bsnm{Intel}, \binits{I.}}:
Intel math kernel library benchmarks (intel r mkl benchmarks).
URL https://software. intel. com/en-us/articles/intel-mkl-benchmarks-suite
\end{botherref}
\endbibitem

\bibitem{Langhammer2018QRFPGAs}
\begin{bchapter}
\bauthor{\bsnm{Langhammer}, \binits{M.}},
\bauthor{\bsnm{Pasca}, \binits{B.}}:
\bctitle{High-performance qr decomposition for fpgas}.
In: \bbtitle{Proceedings of the 2018 ACM/SIGDA International Symposium on
  Field-Programmable Gate Arrays}.
\bsertitle{FPGA '18},
pp. \bfpage{183}--\blpage{188}.
\bpublisher{Association for Computing Machinery},
\blocation{New York, NY, USA}
(\byear{2018}).
\doiurl{10.1145/3174243.3174273}.
\burl{https://doi.org/10.1145/3174243.3174273}
\end{bchapter}
\endbibitem

\bibitem{PUREDatabase2014}
\begin{bchapter}
\bauthor{\bsnm{Stricker}, \binits{R.}},
\bauthor{\bsnm{Müller}, \binits{S.}},
\bauthor{\bsnm{Gross}, \binits{H.-M.}}:
\bctitle{Non-contact video-based pulse rate measurement on a mobile service
  robot}.
In: \bbtitle{IEEE Int. Symp. on Robot and Human Interactive Communication},
pp. \bfpage{1056}--\blpage{62}
(\byear{2014}).
\doiurl{10.1109/ROMAN.2014.6926392}
\end{bchapter}
\endbibitem

\bibitem{COHFACE2017}
\begin{botherref}
\oauthor{\bsnm{Heusch}, \binits{G.}},
\oauthor{\bsnm{Anjos}, \binits{A.}},
\oauthor{\bsnm{Marcel}, \binits{S.}}:
A reproducible study on remote heart rate measurement.
arXiv preprint arXiv:1709.00962
(2017)
\end{botherref}
\endbibitem

\bibitem{UBFCDatabase2019}
\begin{barticle}
\bauthor{\bsnm{Bobbia}, \binits{S.}},
\bauthor{\bsnm{Macwan}, \binits{R.}},
\bauthor{\bsnm{Benezeth}, \binits{Y.}},
\bauthor{\bsnm{Mansouri}, \binits{A.}},
\bauthor{\bsnm{Dubois}, \binits{J.}}:
\batitle{Unsupervised skin tissue segmentation for remote
  photoplethysmography}.
\bjtitle{Pattern Recognition Letters}
\bvolume{124},
\bfpage{82}--\blpage{90}
(\byear{2019})
\end{barticle}
\endbibitem

\bibitem{MAHNOB2021Soleymani}
\begin{barticle}
\bauthor{\bsnm{Soleymani}, \binits{M.}},
\bauthor{\bsnm{Lichtenauer}, \binits{J.}},
\bauthor{\bsnm{Pun}, \binits{T.}},
\bauthor{\bsnm{Pantic}, \binits{M.}}:
\batitle{A multimodal database for affect recognition and implicit tagging}.
\bjtitle{IEEE Transactions on Affective Computing}
\bvolume{3}(\bissue{1}),
\bfpage{42}--\blpage{55}
(\byear{2012}).
\doiurl{10.1109/T-AFFC.2011.25}
\end{barticle}
\endbibitem

\bibitem{Allen2004PPGsBody}
\begin{bchapter}
\bauthor{\bsnm{Allen}, \binits{J.}},
\bauthor{\bsnm{Murray}, \binits{A.}}:
\bctitle{Effects of filtering on multisite photoplethysmography pulse waveform
  characteristics},
pp. \bfpage{485}--\blpage{488}
(\byear{2004}).
\doiurl{10.1109/CIC.2004.1442980}
\end{bchapter}
\endbibitem

\bibitem{EffectsVideoCompression_HR_PPG}
\begin{barticle}
\bauthor{\bsnm{Rapczynski}, \binits{M.}},
\bauthor{\bsnm{Werner}, \binits{P.}},
\bauthor{\bsnm{Al-Hamadi}, \binits{A.}}:
\batitle{Effects of video encoding on camera based heart rate estimation}.
\bjtitle{IEEE Transactions on Biomedical Engineering}
\bvolume{PP},
\bfpage{1}--\blpage{1}
(\byear{2019}).
\doiurl{10.1109/TBME.2019.2904326}
\end{barticle}
\endbibitem

\bibitem{gupta2023radiant}
\begin{bchapter}
\bauthor{\bsnm{Gupta}, \binits{A.K.}},
\bauthor{\bsnm{Kumar}, \binits{R.}},
\bauthor{\bsnm{Birla}, \binits{L.}},
\bauthor{\bsnm{Gupta}, \binits{P.}}:
\bctitle{Radiant: Better rppg estimation using signal embeddings and
  transformer}.
In: \bbtitle{Proceedings of the IEEE/CVF Winter Conference on Applications of
  Computer Vision},
pp. \bfpage{4976}--\blpage{4986}
(\byear{2023})
\end{bchapter}
\endbibitem

\bibitem{Alvarez2023rPPG2Depression}
\begin{botherref}
\oauthor{\bparticle{Álvarez} \bsnm{Casado}, \binits{C.}},
\oauthor{\bsnm{Lage~Cañellas}, \binits{M.}},
\oauthor{\bsnm{Bordallo~López}, \binits{M.}}:
Depression recognition using remote photoplethysmography from facial videos.
IEEE Transactions on Affective Computing,
1--13
(2023).
\doiurl{10.1109/TAFFC.2023.3238641}
\end{botherref}
\endbibitem

\end{thebibliography}


\end{document}